\documentclass[10pt,journal,cspaper,compsoc,twoside]{IEEEtran}

\ifCLASSOPTIONcompsoc
	\usepackage[nocompress]{cite}
  \usepackage[caption=false,font=normalsize,labelfont=sf,textfont=sf]{subfig}
\else
	\usepackage{cite}
  \usepackage[caption=false,font=footnotesize]{subfig}
\fi

\usepackage[pdftex]{graphicx}
\usepackage[cmex10]{amsmath}
\usepackage{amssymb}
\usepackage{array}
\usepackage{fixltx2e}
\usepackage{url}
\usepackage{color}
\usepackage{hyperref}
 \usepackage{multirow}

\graphicspath{{./images/}}
\DeclareGraphicsExtensions{.jpg}

\newcommand{\etal}{et al.\ }

\begin{document}

\title{Ear-to-ear Capture of Facial Intrinsics}

\author{Alassane~Seck,
        William~A.~P.~Smith,~\IEEEmembership{Member,~IEEE,}
        Arnaud~Dessein,\\
        Bernard~Tiddeman,
        Hannah~Dee,
        and~Abhishek~Dutta%
        \IEEEcompsocitemizethanks{\IEEEcompsocthanksitem A. Seck, B. Tiddeman and H. Dee are with the Department of Computer Science, Aberystwyth University, UK.
Email: \{als31,bpt,hmd1\}@aber.ac.uk
\IEEEcompsocthanksitem W. Smith and A. Dessein are with the Department of Computer Science, University of York, UK.
Email: \{william.smith,arnaud.dessein\}@york.ac.uk
\IEEEcompsocthanksitem A. Dutta is with the University of Twente, Netherlands.
Email: a.dutta@utwente.nl
}% <-this % stops a space
\thanks{}}

\markboth{Seck \MakeLowercase{\textit{\etal}}: Ear-to-ear Capture of Facial Intrinsics}%
{Seck \MakeLowercase{\textit{\etal}}: Ear-to-ear Capture of Facial Intrinsics}

\IEEEcompsoctitleabstractindextext{%
\begin{abstract}
We present a practical approach to capturing ear-to-ear face models comprising both 3D meshes and intrinsic textures (i.e. diffuse and specular albedo). Our approach is a hybrid of geometric and photometric methods and requires no geometric calibration. Photometric measurements made in a lightstage are used to estimate view dependent high resolution normal maps. We overcome the problem of having a single photometric viewpoint by capturing in multiple poses. We use uncalibrated multiview stereo to estimate a coarse base mesh to which the photometric views are registered. We propose a novel approach to robustly stitching surface normal and intrinsic texture data into a seamless, complete and highly detailed face model. The resulting relightable models provide photorealistic renderings in any view.
\end{abstract}

\begin{IEEEkeywords}
Diffuse albedo, face capture, multiview stereo, photometric stereo, specular albedo.
\end{IEEEkeywords}}

\maketitle

\IEEEdisplaynotcompsoctitleabstractindextext

\ifCLASSOPTIONpeerreview
	\begin{center} \bfseries EDICS Category: \end{center}
\fi

\IEEEpeerreviewmaketitle

\ifCLASSOPTIONcompsoc
	\noindent\raisebox{2\baselineskip}[0pt][0pt]%
	{\parbox{\columnwidth}{\section{Introduction}\label{sec:introduction}%
	\global\everypar=\everypar}}%
	\vspace{-1\baselineskip}\vspace{-\parskip}\par
\else
	\section{Introduction}\label{sec:introduction}\par
\fi

\IEEEPARstart{M}{easuring} properties of a face that are truly intrinsic (i.e. unrelated to environmental or imaging parameters) is a longstanding goal with applications in a wide range of fields. In graphics, it allows face images to be synthesised in arbitrary illumination conditions \cite{Ma:07}, simulating the properties of any camera. In statistical modelling, it allows the variability in a population of faces to be studied independently of imaging conditions \cite{Blanz:99}. In computer vision, it allows appearance in an image to be predicted for pose and illumination invariant recognition or classification \cite{Basri:03,Blanz:03b}. In psychology, it allows the relative importance of shape and intrinsic texture to the neural representation of faces to be studied \cite{Jiang:09}. Despite these broad and compelling applications, and over a decade of research attention, there remains no satisfactory method for capturing intrinsic properties of a whole face (from ear-to-ear).

By ``intrinsic properties'' we refer specifically to the shape and reflectance properties of a face that give rise to a particular face appearance when illuminated and imaged. Shape is usually represented by a 3D mesh and reflectance properties by 2D parameter maps representing the spatial distribution of reflectance parameters in texture space. In turn, reflectance parameters are determined by the spatial distribution of biophysical parameters such as skin pigmentation and facial hair over the face surface.

Estimating reflectance properties necessitates photometric measurements. However, the state-of-the-art approach \cite{Ma:07,Wilson:10} relies on a view dependent calibration of the orientation of polarising filters, restricting the method to a single viewpoint from which a face is only partially visible. In any single view, parts of the face are either occluded or so foreshortened that their projected resolution is too low to provide useful information. However, for many applications a full face model is required. For example, it has been shown that the ears are an important feature for 3D face modelling \cite{Bustard:10}. Likewise, cropped face models introduce artificial boundaries that make it difficult to use the model as part of a character animation or may disrupt neural processes when used as psychological stimuli.

In this paper we present an approach that enables the intrinsic shape and reflectance properties to be captured over the whole face surface. 

\renewcommand{\arraystretch}{0.0}
\begin{figure*}[t]
\centering
\begin{tabular}{@{}c@{}c@{}c@{}c@{}c@{}}
		{\scriptsize Diffuse albedo} & {\scriptsize Specular albedo} & {\scriptsize Normal map} & {\scriptsize Stitched geometry} & {\scriptsize Rendering} \\
		\includegraphics[width=2cm]{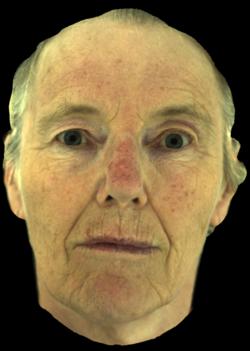} &
		\includegraphics[width=2cm]{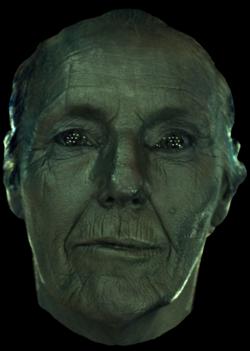} &
		\includegraphics[width=2cm]{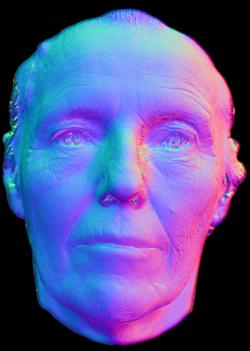} &
		\multirow{3}{*}[75pt]{\includegraphics[height=7.5cm]{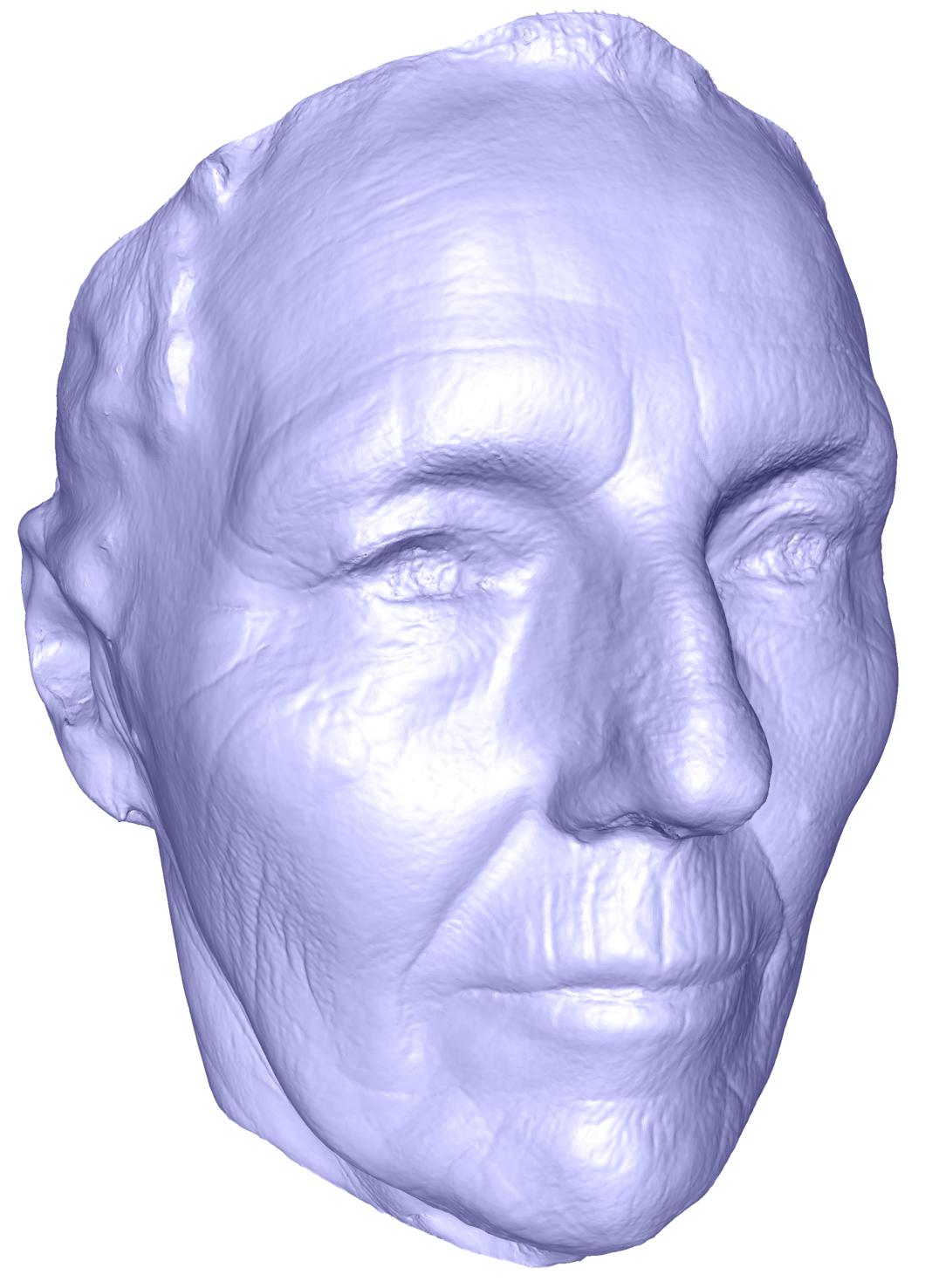}} &
		\multirow{3}{*}[75pt]{\includegraphics[height=7.5cm]{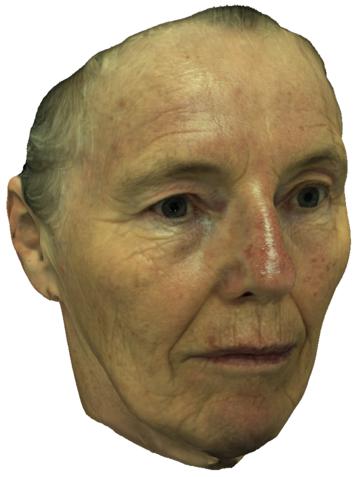}}
		\\
		
		\includegraphics[width=2cm]{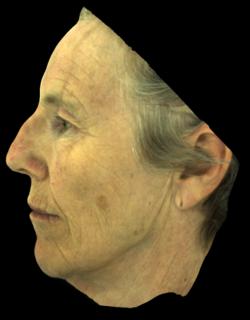} &
		\includegraphics[width=2cm]{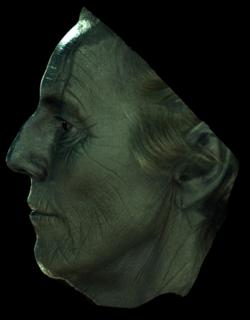} &
		\includegraphics[width=2cm]{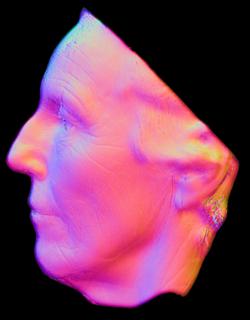} & & 
		\\
		
		\includegraphics[width=2cm]{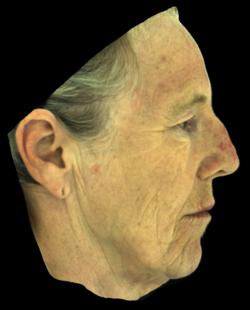} &
		\includegraphics[width=2cm]{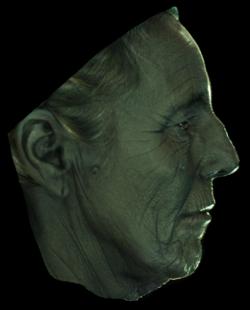} &
		\includegraphics[width=2cm]{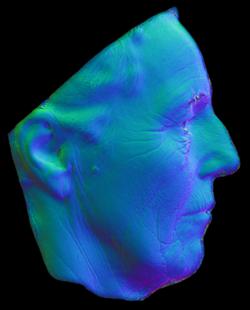} & &

\end{tabular}

\caption{Ear-to-ear face capture: View-dependent photometric observations (left), stitched geometry (middle) and rendering using stitched diffuse/specular albedo and geometry (right).}\label{fig:teaser}

\end{figure*}

\renewcommand{\arraystretch}{1}

\subsection{Related Work}

Existing methods for face shape capture fall broadly into two categories: photometric and geometric. Photometric methods use the intensity of reflected light to infer the orientation and material properties of the face surface. Geometric methods use feature point positions observed from multiple viewpoints to infer the depth of the face surface. The advantage of photometric methods is that they are dense (measurements are made at every pixel and resolution is limited only by the resolution of the camera). Moreover, photometric analysis allows estimation of additional reflectance properties such as diffuse and specular albedo \cite{Ma:07}, surface roughness \cite{Ghosh:09} and index of refraction \cite{Ghosh:10}. This information is essential for rendering or relighting the captured face models. 

However, surface orientation is only a 2.5D shape representation and the estimated normal field must be integrated in order to recover surface depth \cite{zafeiriou.2011} or used to refine a 3D mesh captured using other cues \cite{Nehab:05}. Also, photometric methods are usually much more demanding in data capture terms and also in their requirement for controlled conditions.

Geometric methods on the other hand allow instantaneous capture of face geometry but do not allow estimation of reflectance properties, providing only a fixed texture map. Below, we review existing methods for photometric and geometric face capture as well as hybrid methods.

{\bf Photometric methods.}  Photometric shape estimation has a long history in computer vision \cite{Woodham:80}. Recently, there has been a resurgent interest in applying photometric methods to the problem of face capture \cite{Ma:07,Wilson:10,Ghosh:09,Ghosh:10}. In particular, the spherical gradient-based photometric stereo method of Ma \etal \cite{Ma:07} allowed finescale facial features such as skin pores and wrinkles to be recovered with enhanced accuracy and robustness in comparison to traditional point source methods. This was extended to realtime performance capture by Wilson \etal \cite{Wilson:10} who showed how a certain sequence of illumination conditions allowed for temporal upsampling of the photometric shape estimates.

A lightstage uses polarisation to separate specular and diffuse reflectance. This is most easily achieved by placing a linear polarising filter in front of each point source. The filter is oriented such that, once reflected specularly from the face, the plane of polarisation is the same for all sources. Unfortunately, the required orientation is viewpoint dependent: a given calibration only separates diffuse and specular reflection for two (antipodal) viewing directions. Hence, there is no straightforward way to perform multiview photometric analysis using a lightstage. Going further, when using a single photometric view such as in the approach of Ma \etal \cite{Ma:07}, the albedo maps are corrupted by ``ambient occlusion'' and inter-reflection effects meaning shape and material properties are not fully separated.

{\bf Geometric methods.} Multiview shape estimation methods such as binocular stereo, structure-from-motion and multiview stereo have been applied quite successfully to the problem of face shape estimation. The key problem in this context is establishing correspondence between views over apparently-featureless regions of the face such as the cheeks and forehead. One solution to this problem is to either paint \cite{Furukawa:08} or project \cite{zhang2013viewport} a pattern onto the face that provides matchable features. An alternative passive approach is to use very high resolution images in which fine scale features such as freckles, wrinkles and skin pores are resolved. These provide ideal features for robust matching. Finally, methods such as shape-from-silhouette do not rely on feature matching and can hence be applied to faces \cite{Moghaddam:03} even when small features are not visible.

The state-of-the-art approach in geometric face capture is due to Beeler \etal \cite{Beeler:10}. Since their method is reliant on a very accurate geometric calibration, they propose a novel  calibration process based on a spherical calibration target. Shape estimation then proceeds in two steps. First a base mesh is obtained using a multiview stereo approach. Next, detail is embossed onto the mesh using a shading-based heuristic. Whilst the resulting meshes contain convincing detail, the fine scale shape detail is not accurate since it is hallucinated from a texture cue rather than satisfying any meaningful geometric or photometric constraint.

In general, any multiview method that relies on accurate intrinsic and extrinsic calibration is highly restrictive. The camera focus must be fixed between calibration and face capture. This is particularly problematic if such a setup is to be integrated with a photometric system. For example, in a lightstage the amount of light received by the camera is limited by the polarising filters on illuminants and camera. This usually means that a relatively large aperture is used, leading to a reduced depth of field. In such a case, focus is very sensitive and it is unlikely that a single focus would suffice for both calibration and capture.

Of course, purely geometric methods can only recover one intrinsic property of faces: shape. Texture maps are nothing other than a photograph of the face under a particular set of environmental conditions. Hence, the texture obtained using multiview stereo is useless for relighting. Worse, since appearance is view-dependent (the position of specularities changes with viewing direction), no one single appearance can explain the set of multiview images. 

{\bf Hybrid methods} There have been a number of attempts to combine photometric and geometric methods for face or object capture. This is largely motivated by the fact that their advantages complement the weaknesses of the alternate approach.

Nehab \etal \cite{Nehab:05} proposed an efficient approach for combining estimated surface normals and surfaces (in the form of a depth map or mesh). Their approach is particularly applicable to surface normals estimated using photometric methods which are likely to contain low frequency bias. This low frequency bias is removed by the base mesh which in return is refined by the accurate high frequency detail present in the photometric surface normal. Their approach is based on a linear approximation to the underlying objective of minimising angular error between target normals and those of the final surface.

Wu \etal \cite{wu2011fusing} propose an approach that combines multi-illumination MVS and uncalibrated photometric stereo methods. They recover depth maps from multi-view and multi-illumination video sequences, then merge these to a watertight mesh using multi-illumination photo-consistency constraints. The recovered mesh is further refined with photometric surface normals measured under uncalibrated (unknown) illumination conditions. The proposed uncalibrated photometric stereo technique consist in an iterative estimation of both the surface normals and the illumination conditions simultaneously. Uncalibrated photometric stereo techniques are also used by Park \etal \cite{park.2013}. Although to bypass the problem of merging multiple views photometric data, they compute the photometric normals directly in the 2D parameter domain. 
 
The closest previous work to what we propose in this paper is due to Ghosh \etal \cite{ghosh2011multiview}. They capture multiple view photometric data simultaneously in a lightstage. In order to overcome the view dependency of the polariser orientation calibration, they make an empirical observation. Namely, using two illumination fields with locally orthogonal patterns of polarisation (i.e. filters aligned to lines of latitude or longitude) allows approximate specular/diffuse separation from any view close to the equator.
Unfortunately, this approximation means that their approach does not recover truly intrinsic properties and therefore does not fulfil our goals.
Specular and diffuse reflectance is not fully separated meaning that both normal maps and specular/diffuse albedo maps are corrupted.
In addition, they propose an empirical process for ``Fresnel compensation'' of the specular albedo maps. As we discuss in Section \ref{fresnel_gain}, this is based on an assumption of constant specular reflectance properties, and neglects rough surface effects. This further reduces the utility of the specular albedo as an intrinsic quantity.

\subsection{Contributions}

\begin{figure*}[t]
\centering
\includegraphics[width=16cm]{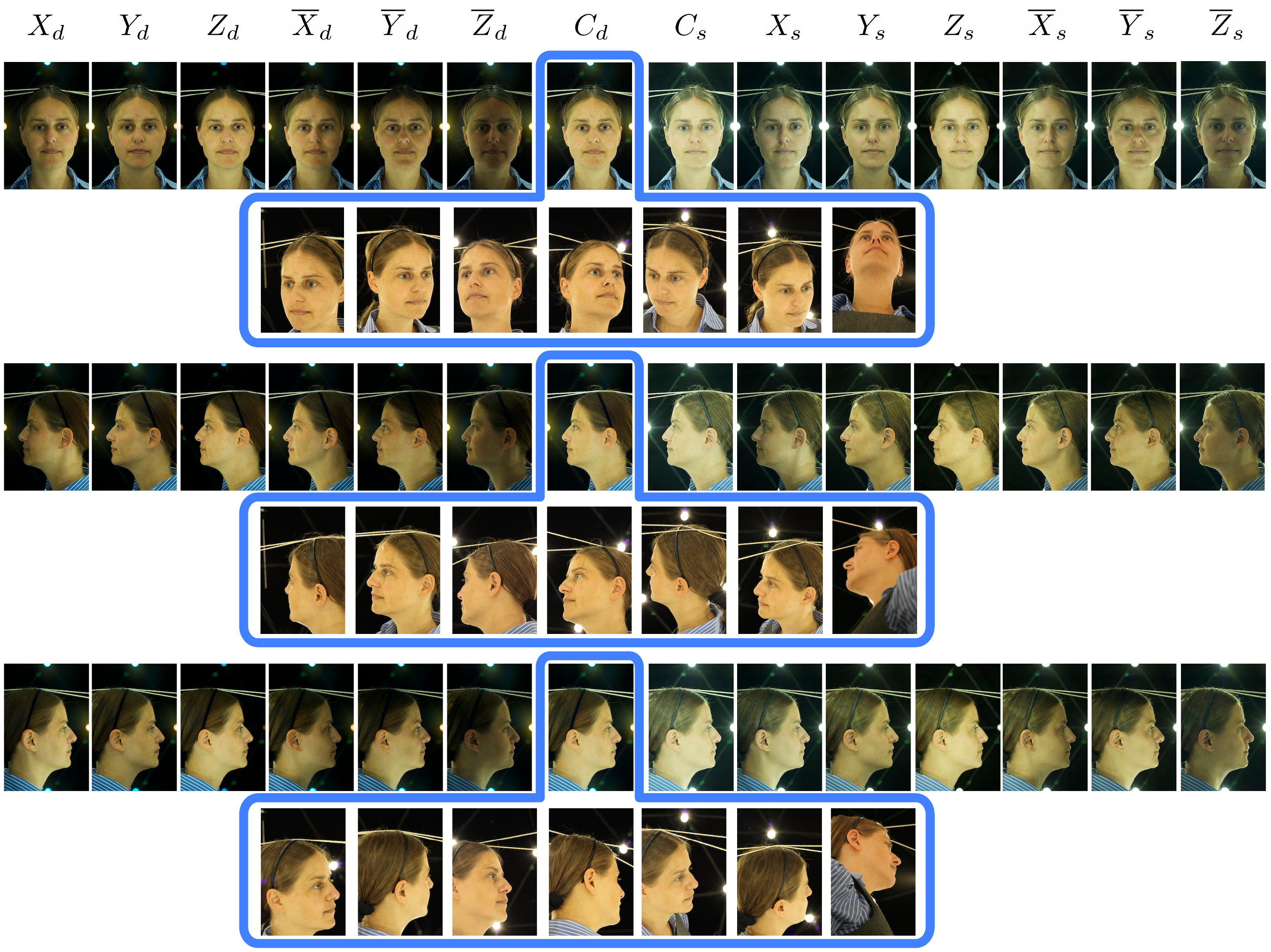} 
\caption{The set of 63 images used in our pipeline. We obtain 24 different viewpoints of the face (boxed blue images) by capturing 8 different views with the subject in a frontal (rows 1 and 2), left profile (rows 3 and 4) and right profile (rows 5 and 6). Rows 1, 3 and 5 show the sequence of images captured for the photometric viewpoint. Rows 2, 4 and 6 show the multiview images. Images within a blue box are captured simultaneously. The remainder are captured in sequence from left to right.}
\label{fig:image_set}
\end{figure*}

Our proposed approach enables photometric measurements to be made over the whole face. In contrast to Ghosh \etal \cite{ghosh2011multiview}, we do not attempt to capture multiple photometric views simultaneously. Instead, we capture from a single view in multiple poses and tackle the resulting stitching problem.
We begin by proposing a novel capture pipeline that uses uncalibrated multiview stereo to register photometric views to a base mesh.  We then estimate single view surface normal and reflectance information using the spherical gradient photometric stereo approach of Wilson \etal \cite{wilson2010temporal}. In this single view setting, we propose two improvements. First, we use the base mesh to remove the view-dependent low frequency bias present in the estimated surface normals. Second, we provide a photometric alignment method that avoids the iterative approach of Wilson \etal \cite{wilson2010temporal}. Finally, in order to merge the separate views we present a robust approach to stitching photometric information across multiple views. This unified approach allows us to stitch both textures and shape as a solution of a screened Poisson equation. In the case of stitching shape, the problem is closely related to Laplacian mesh editing.

Our approach requires no geometric calibration, no inverse rendering, only consumer hardware and is fast (the whole capture process comprises three sequences lasting around 3 seconds each). Yet, the quality of the estimated shape is comparable to state-of-the-art methods that require much more careful calibration, e.g. \cite{Beeler:10}. We have the additional advantage that our method also estimates intrinsic diffuse and specular reflectance maps meaning our models are relightable.

\section{Pipeline}
The polarisation properties of light have been widely used as a cue to study surface shape and reflectance properties. One of the best known effects is that specular reflectance from a dielectric material preserves the plane of polarisation of linearly polarised incident light whereas the diffuse reflectance loses it. This allows separation of specular and diffuse reflectance using a cross-polarisation technique \cite{ma2007rapid}. However, while calibrating this technique is relatively straightforward it is, unfortunately, view dependent.

We overcome this problem by capturing a face multiple times in different poses relative to the calibrated viewpoint, e.g. frontal and two profile views. Together, these three photometric views provide full ear-to-ear coverage of the face. We augment the photometric camera with additional cameras providing multiview, single-shot images captured in sync with a reference frame of the photometric sequence (the diffuse constant image). We position these additional cameras to provide overlapping coverage of the face. Since we do not rely on a fixed calibration, their exact positioning is unimportant and we allow the cameras to autofocus between captures. In our setup, we use 7 such cameras in addition to the photometric view giving a total of 8 simultaneous views. Since we repeat the capture three times, we have 24 effective views. A complete dataset for a face is shown in Figure \ref{fig:image_set}.

In order to merge these views and to provide a rough base mesh, we perform a multiview reconstruction using all 24 views. Solving this uncalibrated multiview reconstruction problem provides both the base mesh and also intrinsic and extrinsic camera parameters for the three photometric views. These form the input to our stitching process. Note that since the three photometric views are not acquired simultaneously, there is likely to be non-rigid deformation of the face between these views. For this reason, in Section \ref{sec:stitching} we propose a robust algorithm for stitching the views without blurring potentially misaligned features.

Our complete pipeline is summarised as follows:
\begin{enumerate}
\item {\bf Uncalibrated multiview stereo}: We commence by applying structure-from-motion followed by dense multiview stereo to all 24 viewpoints. 
\item {\bf Photometric capture}: For the three photometric viewpoints, we capture 14 image spherical gradient illumination sequences. These comprise the 7 gradient conditions with crossed and parallel polarised filter orientations on the camera.
\item {\bf Per-view alignment and bias removal}: For each photometric viewpoint, we compensate for subject motion using the photometric alignment technique described in Section \ref{photo_align} and estimate diffuse and specular surface normal maps. We perform bias removal for each view, accounting for the pose-dependency of light source discretisation on the estimated normals.
\item {\bf Stitching photometric views}: Finally, we stitch the diffuse and specular albedo and surface normals onto the base mesh. The surface normal stitching is done in the mesh domain so that the detail is transferred to the vertices simultaneously with stitching the normals.
\end{enumerate}
Step 1 is now a well studied problem and with high resolution face images, satisfactory results can be obtained using existing methods such as the Bundler SFM tool \cite{Snavely:2006} and PMVS for multiview stereo \cite{Furukawa:2009}. We use the commercial tool Agisoft Photoscan \footnote{www.agisoft.com}. For step 2 we use an opto-electrical polarising filter to allow diffuse/specular separation without the need for mechanical filter rotation. Such filters form part of active 3D projection systems and are available cheaply. Optionally, for reasons of efficiency it may be desirable to decimate the final mesh and store texture and shape detail in 2D maps. This can be done as a post-processing step to our pipeline using any existing surface parameterisation and decimation algorithms. 

\section{Spherical Gradient Photometric Stereo}

Spherical Gradient Photometric Stereo was introduced by Ma \etal \cite{Ma:07} and refined by Wilson \etal \cite{Wilson:10}. The idea amounts to something very simple: estimate the first moment (centre of mass) of the reflectance lobe at a point by illuminating that point with a linear spherical gradient. For a Lambertian surface, this direction coincides with the surface normal and, for a specular surface, with the reflection direction (from which the surface normal can be calculated).

\subsection{The Lambertian case}

Let $X_d$, $Y_d$, $Z_d$ and $C_d$ respectively the measured Lambertian radiances under the $X-$gradient, $Y-$gradient, $Z-$gradient and constant illuminations, \cite{Ma:07} established the relation between the surface normal $n^d=(n_x^d,n_y^d,n_z^d)$ and the measured Lambertian radiance as follows:
\begin{eqnarray}
n_x^d & = & \frac{1}{N_{d}}\left(\frac{X_d}{C_d} - \frac{1}{2}\right) \nonumber \\
\label{eq:ppalsi_eq9}
n_y^d & = & \frac{1}{N_{d}}\left(\frac{Y_d}{C_d} - \frac{1}{2}\right) \nonumber \\
\label{eq:ppalsi_eq10}
n_z^d & = & \frac{1}{N_{d}}\left(\frac{Z_d}{C_d} - \frac{1}{2}\right) 
\label{eq:ma2007_ratio_eq},
\end{eqnarray}
where, $N_{d}$ is a normalizing constant. 

\subsection{The specular case}
In the specular case, \cite{Ma:07} show that it is easier to estimate, from the measured specular radiances, the specular reflection vector than the surface normal directly. If $X_s$, $Y_s$, $Z_s$ and $C_s$ denote respectively the measured specular radiances under  $X-$gradient, $Y-$gradient, $Z-$gradient and constant illuminations, the specular reflection vector $u=(u_x,u_y,u_z)$ is given by:
\begin{eqnarray}
u_x & = & \frac{1}{N_{s}}\left(X_s - \frac{1}{2}C_s\right) \nonumber \\
u_y & = & \frac{1}{N_{s}}\left(Y_s - \frac{1}{2}C_s\right) \nonumber \\
u_z & = & \frac{1}{N_{s}}\left(Z_s - \frac{1}{2}C_s\right) ,
\end{eqnarray}
where, $N_{s}$ is a normalizing constant. 

As the surface normal corresponds to the direction half-way between the view vector $v$ (which is $v=[0\ 0\ 1]^T$ in our case) and its specular reflection $u$, it can be obtained by:
\begin{equation}
{n}^s = \frac{u+v}{\|u+v\|}.
\end{equation} 

\subsection{Complement Gradient Illumination}

Wilson \etal \cite{Wilson:10} proposed an improved method for calculating the surface normals from Spherical Gradient Illumination. The authors exploit spherical gradient images obtained under complementary lighting conditions, i.e. for which the lighting coordinate system is reversed. Thus, in addition to the four gradient images $X$,$Y$, $Z$ and $C$ proposed by Ma \etal \cite{Ma:07}, they capture three others $\bar{X}$,$\bar{Y}$ and $\bar{Z}$ such that:
\begin{equation}
X +\bar{X} = Y + \bar{Y} = Z + \bar{Z} = C. 
\label{eq:complement}
\end{equation}   
From \ref{eq:complement}, \ref{eq:ma2007_ratio_eq} and a re-normalization, they obtain:
\begin{equation}
n = \frac{\left[ X - \bar{X}, Y - \bar{Y} , Z - \bar{Z} \right]^T}{\| \left[ X - \bar{X}, Y - \bar{Y} , Z - \bar{Z} \right]^T \|}.
\label{eq:complement_normal}
\end{equation}
This method is proven to improve the quality of the calculated normals and is more robust than the original method of Ma \etal \cite{Ma:07}. This is explained by the fact that the dark regions in one gradient image are likely to be well lit in the complement image, hence improving signal to noise ratio.

\section{Photometric Alignment}\label{photo_align}

Since spherical gradient photometric stereo requires a set of images to be captured in series, the images within a sequence may not be in perfect alignment due to subject motion. In the context of estimating fine scale shape, these small misalignments lead to a blurring of detail. Since inter-frame motion is likely to be very small (perhaps sub-pixel) and visibility is unlikely to change between views, the obvious solution is to use optical flow to align each image to a reference frame. However, due to illumination changes in each frame the usual brightness constancy constraint does not apply (we expect the brightness of a given point on the face to vary dramatically as illumination changes).

Wilson \etal \cite{Wilson:10} overcame this problem by exploiting a property of the complement images. Assuming no motion, the sum of a gradient image and its complement are equal to the constant image. Hence, an alternative brightness constancy constraint can be written down. For example, for the $x$-gradient images:
\begin{equation}
C(x,y)=X(x+\Delta x_1,y+\Delta y_1)+\bar{X}(x+\Delta x_2,y+\Delta y_2).
\end{equation}
This involves solving for the optical flow vectors for both gradient and complement images in one go. Wilson \etal \cite{Wilson:10} propose an iterative approach to this problem where they initially compute the flow from $X$ to $C-\bar{X}$ followed by the flow from $\bar{X}$ to $C-w(X)$, where $w$ is the warp computed at the previous step. It is proposed that iterating these two steps converges towards the correct flow for both images.

A weakness of their approach is that $C-\bar{X}$ is not necessarily a good target for warping. As they are not aligned, taking their difference leads to a blurring of features to which $X$ is unlikely to be satisfactorily warped. In an extreme case, it can be shown that this method can fail completely. 

We propose an alternative that is both more efficient and more robust. We note that changing the spherical illumination pattern affects only intensity and not colour. Thus we use color space transformations to extract intensity-free information from images with different illumination condition. Hue-Saturation-Value(HSV) and normalized-RGB color spaces are known to be efficient ways of separating intrinsic color from shading related-intensity \cite{Mallick2005}. For an image $I$, we combine the Hue component of the HSV space with normalized-RGB to produce an illumination-independent image $I_{color}$:

\begin{equation}
I_{color}=\frac{1}{2}\left\{ \textrm{hue}(I)+\frac{I}{\sqrt{I_R^2 + I_G^2 + I_B^2}}\right\}
\end{equation}

\begin{figure}[t]
\centering
\begin{tabular}{cc}
X-Gradient & Constant \\
\includegraphics[height=4cm]{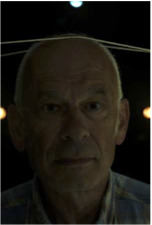} & 
\includegraphics[height=4cm]{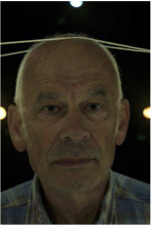} \\

\includegraphics[height=4cm]{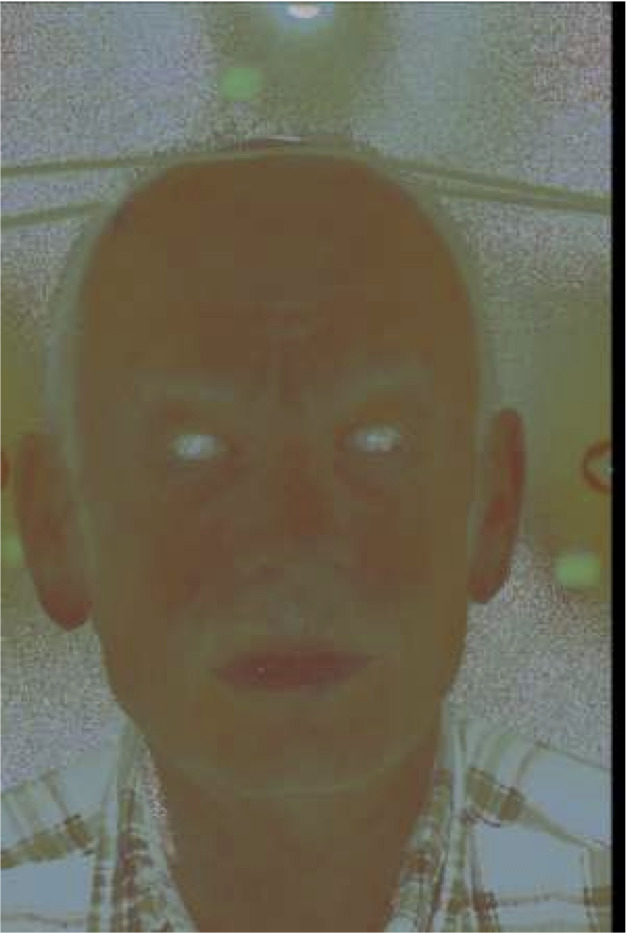}&
\includegraphics[height=4cm]{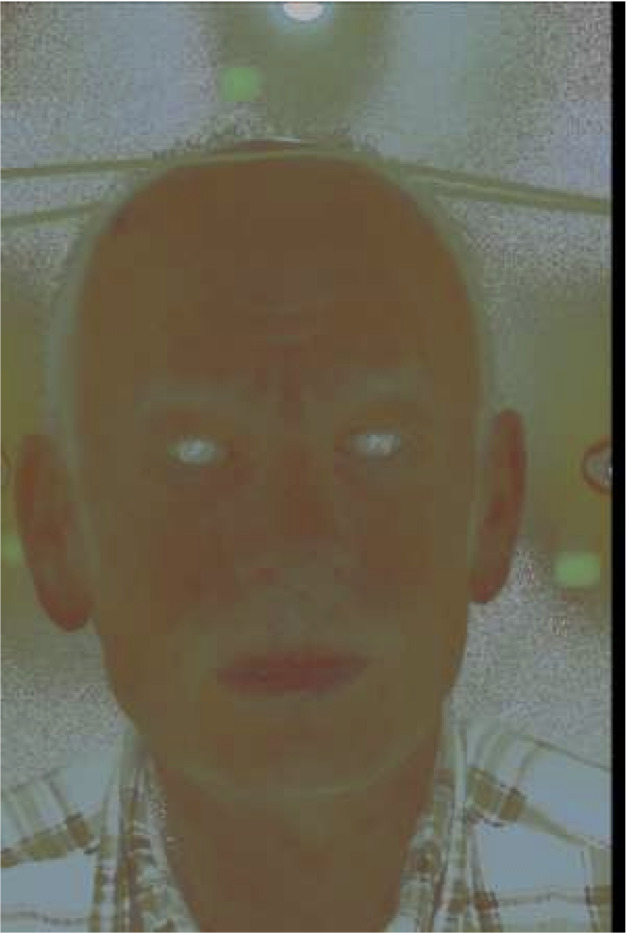} \\

\end{tabular}
\caption{Illumination-independent images for Photometric Alignment}
\label{Illumination-independent-images}
\end{figure}

Figure \ref{Illumination-independent-images} shows two images in different spherical gradient lighting patterns (X-gradient and Constant) and the corresponding illumination-independent images. 

\begin{figure}[t]
\centering
\begin{tabular}{ccc}
\includegraphics[height=3cm]{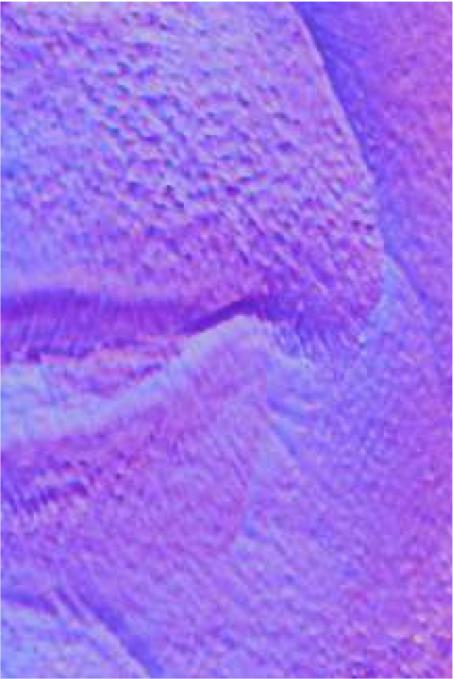} & 
\includegraphics[height=3cm]{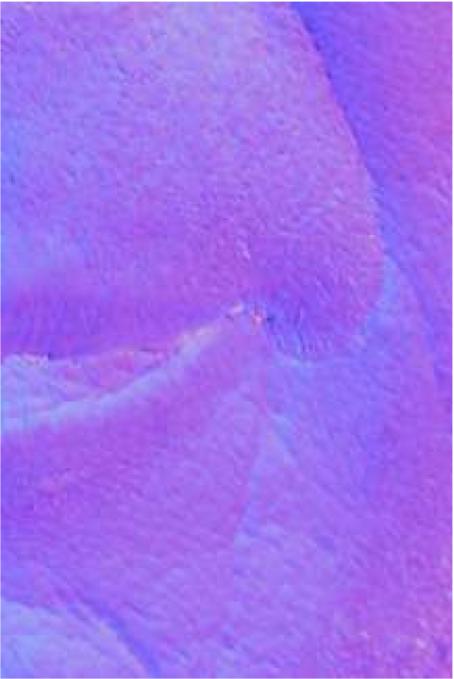} &
\includegraphics[height=3cm]{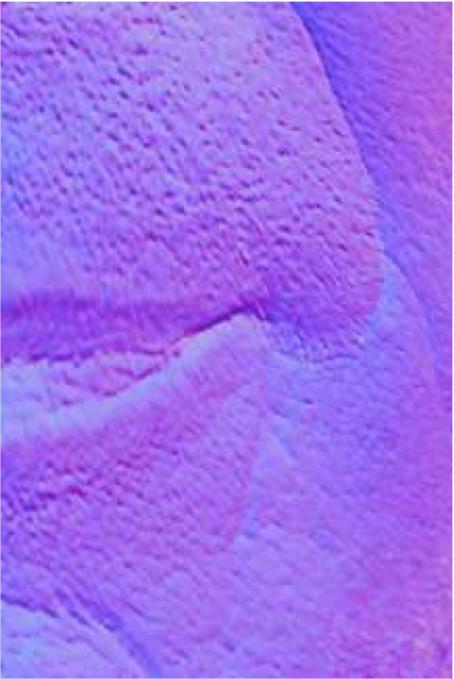} \\
(a) & (b) & (c)\\

\end{tabular}
\caption{Normal-maps obtained with different alignment strategies. (a) Wilson's method after 3 iterations. (b) Our method after 0 iteration (only initialization). (c) Our method after 1 iteration}
\label{Normal-maps-obtained-with-different-alignment-strategies}
\end{figure}

However, while allowing good alignments on global shapes, the color transformation tends to smooth out fine details which can lead to local misalignments. These are more significant as the motion is not rigid. We correct this by employing our method to initialize Wilson's method: we use the flow between $C_{color}$ and $\bar{X}_{color}$ to align $C$ and $\bar{X}$ before computing $C-\bar{X}$. In practice, our experiments show that only one iteration after initialization is enough to get very good alignments. Figure \ref{Normal-maps-obtained-with-different-alignment-strategies} compares normal-maps obtained when photometric images are aligned with 3 iterations of Wilson's method (a) and only 1 iteration when initialized with our method (b).

\subsection{Bias Removal}

The surface normals estimated using spherical gradient photometric stereo are subject to low frequency bias caused by a number of factors. Perhaps most significant is the discretisation of the illumination environment. The analysis above is based on the assumption that the gradient illumination is a continuous field of illumination. In practice, we use 41 LEDs distributed over a geodesic dome.
Second, the method assumes that the light sources are distant and so attenuation effects are constant over the face surface. This is not the case as a head is fairly large (approximately 15cm wide) relative to the size of the geodesic dome (diameter 1.8m). Third, there is also an assumption of no occlusions. Hence, for concave regions of the face, the surface normal is biased towards the unoccluded directions. Other sources of noise such as errors in the light source positions, imperfect specular/diffuse separation and camera nonlinearities further bias the estimated normals.

Many of these effects are pose dependent. For example, the light source positions relative to the face (and hence the discretisation effects) change as the pose of the face changes. For this reason, we perform low frequency bias removal for each photometric view independently prior to stitching. To do so, we use the base mesh provided by multiview stereo to project a depth map into each photometric view. We then combine the high frequency components of the photometric normals with the low frequency components of the depth map normals using the method of Nehab \etal \cite{Nehab:05}. Finally, we transform the corrected normals into world coordinates by applying a rotation based on the extrinsic parameters estimated for that view by structure from motion.

\section{Stitching photometric views}\label{sec:stitching}

In this section, we describe a method for seamlessly stitching the intrinsic textures and normal maps from each photometric view onto the base mesh obtained with multiview stereo. This is a non-trivial problem.

The constraints of linear polarisation necessitate that the three photometric image sets are taken at different times (the subjects rotate themselves to allow capture of one frontal and two profile views). Hence, the face is likely to have changed shape between views, meaning that there is no single correct shape and that correspondence between images and mesh is imperfect. Moreover, certain reflectance effects introduce a view-dependency on the intrinsic textures. For example, Fresnel gain means that specular albedo is unreliable close to occluding boundary (see Figure~\ref{fig:teaser}). Applying a baseline texture stitching algorithm (such as back-projection and averaging) to such data leads to blurring of misaligned features, visible seams between textures taken from different views and inclusion of unwanted specular effects. In addition, we are not aware of any previous work that tackles the problem of stitching normal maps.

To address these problems, we propose a unified approach that allows us to stitch both intrinsic textures and shape. Our approach is based on Poisson blending using non-conservative guidance fields. The guidance fields are either in the form of texture gradients or photometric surface normals. Our approach uses overlapping patches. Within a patch, the guidance field is taken from the single best view (the one with least average viewing angle). In overlap regions, we make per-vertex (for shape) or per-triangle (for texture) selections.
The majority of texture stitching algorithms are vertex- or face-based strategies with additional heuristics for robustness. We expect a patch-based approach to improve robustness by allowing selection criteria to be aggregated over a patch. Also, since a patch is taken from a single view, there will be no blending artefacts within a patch while the patch overlaps provide a means to blend between textures taken from different views.

Our stitching pipeline is as follows. We begin by sampling the photometric observations onto the base mesh provided by multiview stereo.
For each view, we determine the set of visible vertices on the mesh. Each of the intrinsic textures (diffuse/specular albedo and normal map) are then sampled onto the mesh by back-projection for visible vertices after bilinear interpolation within the pixel grid. Additionally, the viewing angles for each face and vertex are computed as part of the process and stored for later use. We then segment the base mesh into overlapping, uniformly sized patches. Finally, we perform stitching using two techniques based on Poisson blending.

\subsection{Mesh segmentation}

We achieve mesh segmentation with a classical farthest-point strategy~\cite{Peyre2006}, enhanced with an original patch growing scheme to form an overlapping structure. We consider a triangular base shape mesh $\mathcal{M}$ and assume it describes a 2D manifold $S$. The connectivity is given by a simplicial complex $\mathcal{K}$ whose elements are vertices $\{i\}$, edges $\{i, j\}$ or faces $\{i, j, k\}$, with indices $i, j, k \in [1 \, .. \, N]$, where $N$ is the number of vertices. We write a vertex $\{i\}$ as $i$ for simplicity.

We first select vertices iteratively by adding a new sample one at a time. Our mesh is equipped with a geodesic distance map $D$.
Denoting by $D_l(i)$ the geodesic distance map to the first $l$ selected samples, we select sample $i_{l + 1}^\star$ as the vertex that maximizes $D_l(i)$. The distance map $D_{l + 1}(i)$ can simply be updated as the minimum between $D_l(i)$ and $D(i, i_{l + 1}^\star)$. We continue this process until a desired number $M$ of vertices have been sampled.

Patches $\mathcal{P}_1, \dotsc, \mathcal{P}_M$ are then obtained via the geodesic Voronoi tessellation based on the samples. The segmentation thus defines a dual graph $\mathcal{G} = (\mathcal{V}, \mathcal{E})$, where $\mathcal{V} = [1 \, .. \, M]$, and $(m, n) \in \mathcal{E}$ if $\mathcal{P}_m$ and $\mathcal{P}_n$ are neighbors, i.e., are connected by an edge $\{i, j\} \in \mathcal{K}$. To grow a patch $\mathcal{P}_m$, we consider separately each of its neighbor patches $\mathcal{P}_n$ with $(m, n) \in \mathcal{E}$, and define thresholds $d_{mn}$ as follows:
\begin{equation}
d_{mn} = \sigma \times D(i_m^\star, i_n^\star) \enspace,
\end{equation}
where $\sigma \geq 0$ is set by the user and can be seen as an overlap ratio or factor, and the geodesic distance $D$ is restricted to the union $\mathcal{P}_m \cup \mathcal{P}_n$ of the reference patch and considered neighbor. The overlap $\mathcal{O}_{mn}$ of $\mathcal{P}_m$ onto $\mathcal{P}_n$ is then constructed by geodesic projections:
\begin{equation}
\mathcal{O}_{mn} = \left\{i \in \mathcal{P}_n \colon \min_{j \in \mathcal{P}_m} D(i, j) \leq d_{mn}\right\} \enspace.
\end{equation}
A given grown patch $\mathcal{Q}_m$ is eventually constructed by concatenation of the reference patch $\mathcal{P}_m$ with the respective overlaps:
\begin{equation}
\mathcal{Q}_m = \mathcal{P}_m \cup \bigcup_{n | (m, n) \in \mathcal{E}} \!\!\!\!\!\! \mathcal{O}_{mn} \enspace.
\end{equation}

\begin{figure}[t]
\centering
\begin{tabular}{@{}c@{}c@{}}
\includegraphics[height=5cm]{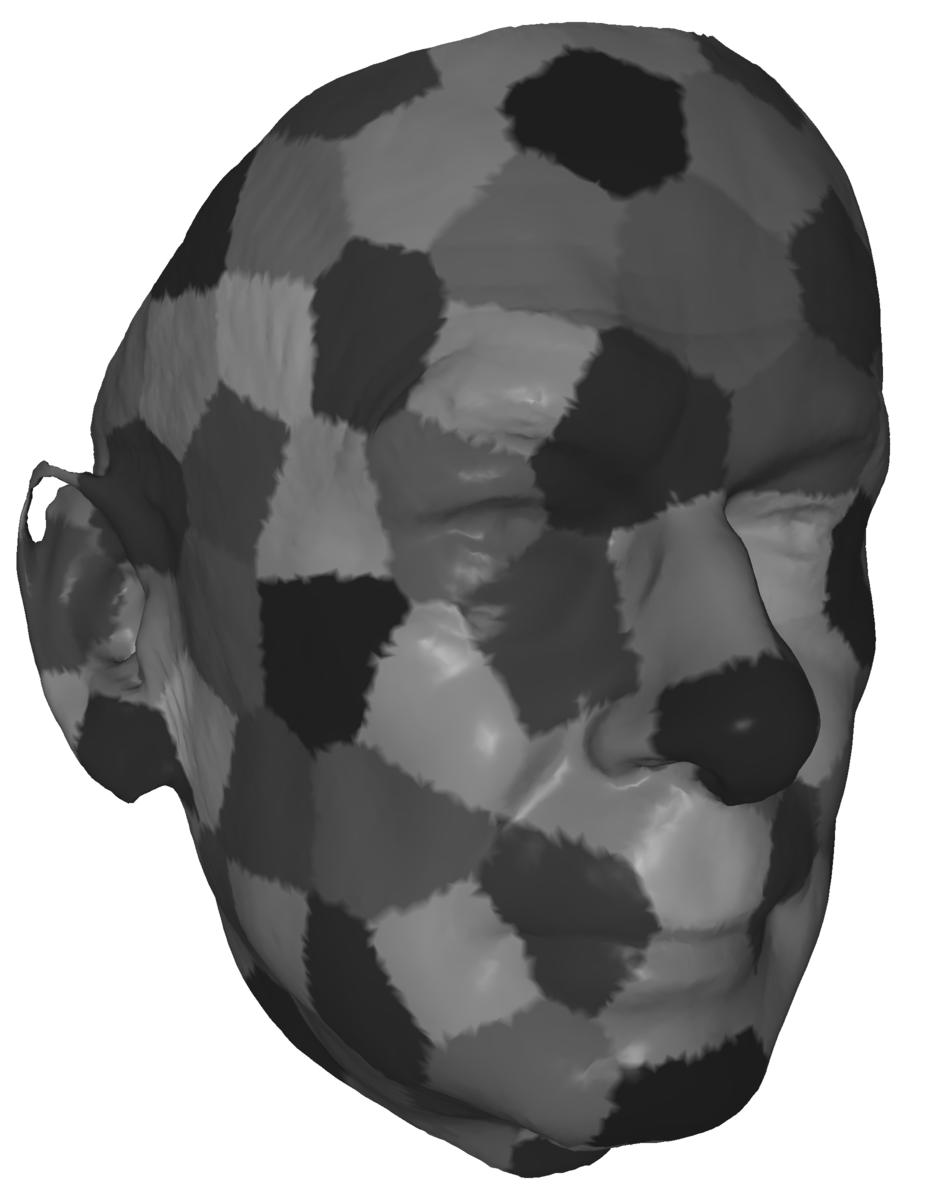} &
\includegraphics[height=5cm]{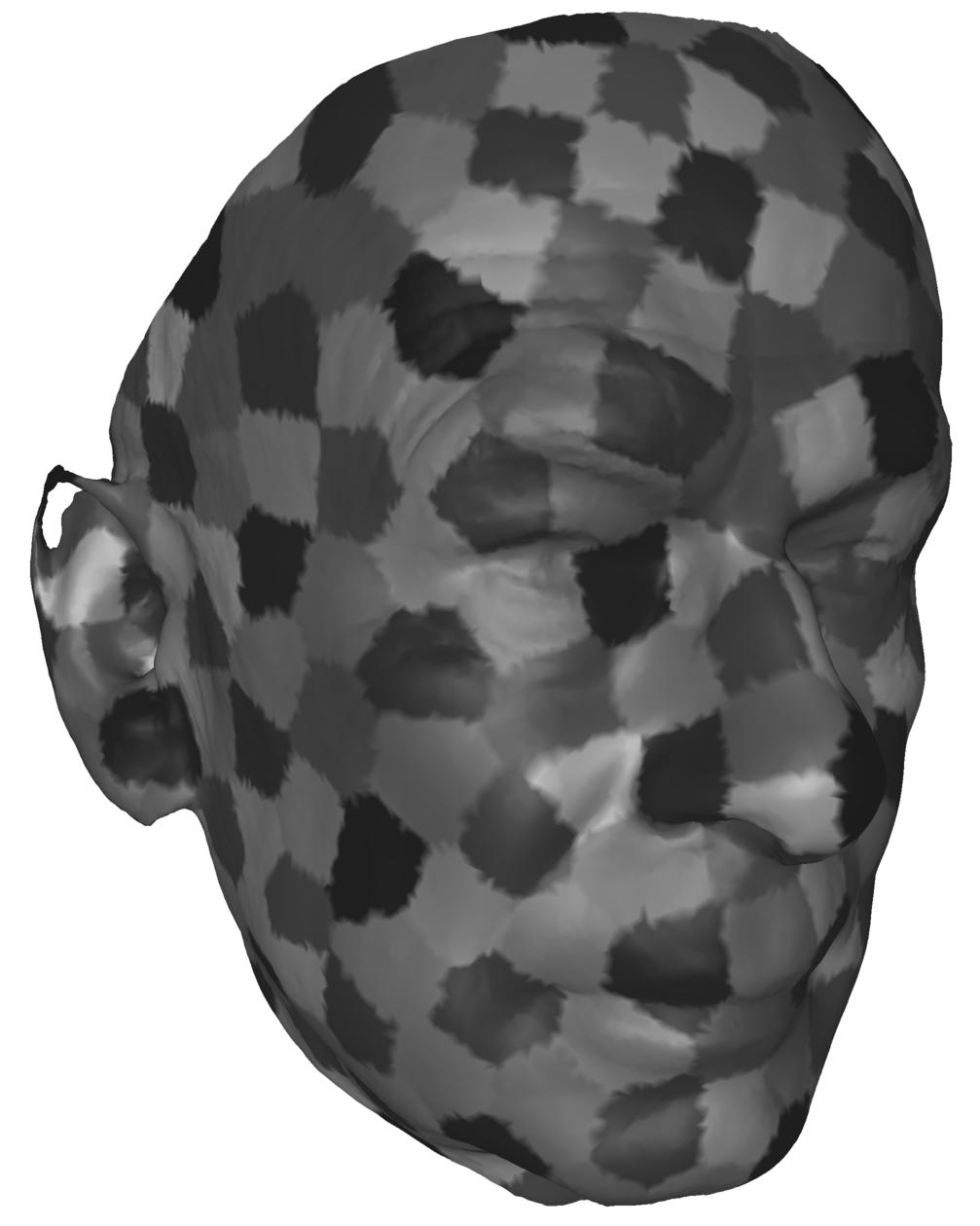}  \\
(a) & (b) \\
\end{tabular}
\caption{Mesh segmentation with different sampling vertices number (left:100; right:400) }
\label{fig:segmenting}
\end{figure}

\subsection{Poisson Blending}

Blending in the gradient domain via solution of a Poisson equation was first proposed by P{\'e}rez \etal \cite{Perez:03} for 2D images. The motivation is that second-order variations in texture are the most significant perceptually whereas low-frequency variations have a barely noticeable effect. The same argument can be made for texture and geometry on a mesh. The approach allows us to avoid visible seams where texture or geometry from different views are inconsistent.

Hence, the idea is to form a guidance field of texture gradients {\bf v} selected from source images and then solve for the texture $f$ whose gradients best match the guidance field:
\begin{equation}
\min_f \iint_{\Omega} {\|\nabla f - {\bf v}\|}^2 dA
\end{equation}
This minimisation problem can be solved by solving the Poisson equation:
\begin{equation}
\Delta f = \nabla\cdot {\bf v},\label{eqn:poisson}
\end{equation}
where $\Delta$ is the Laplace operator and $\nabla\cdot$ is the divergence operator. For non-conservative guidance fields, an exact solution is not possible so Poisson's equation is usually solved in a least squares sense. In our case, the function $f$ is defined over the mesh surface so $\Delta$ is the Laplace-Beltrami operator.
In the case of stitching shape, $f$ becomes the mesh coordinate function.

\subsection{Discrete differential operators}

In order to solve a Poisson equation over a triangle mesh, we need to define discrete counterparts to the Laplace and divergence operators. Since we rely on discrete differential operators on the mesh surface, our approach completely preserves conservative vector fields compared to extrinsic 3D finite elements in~\cite{Chuang2009}. This makes our approach more natural from a theoretical perspective, even if non-conservative fields are rather formed in practice.

A discrete vector field $V$ is a piecewise constant vector function defined for each triangle $T_l$ by a coplanar vector $\mathrm{v}_l$. A discrete potential field is a piecewise linear function $\phi(s) = \sum_{i \in \mathcal{K}} \phi_i B_i(s)$ on the mesh surface, where $B_i$ is the piecewise linear basis function valued $1$ at vertex $i$ and $0$ at other vertices, and $\phi_i$ specifies the value of $\phi$ at vertex $i$. The discrete gradient of $\phi$ for triangle $T_l$ is $\nabla \! \phi_l = \sum_{i \in \mathcal{K}} \phi_i \nabla \! B_{il}$, where $\nabla \! B_{il}$ is the gradient of $B_i$ within $T_l$. The divergence of $V$ at vertex $i$ is $\mathrm{div} \, V \! (i) = \sum_{T_l \in \mathcal{K}_i} |T_l| \, \nabla \! B_{il}^\top \, \mathbf{v}_l$, where $\mathcal{K}_i$ is the set of triangles sharing vertex $i$ and $|T_l|$ is the area of triangle $T_l$. Writing Poisson's equation $\mathrm{div} \, \nabla \! \phi = \mathrm{div} \, V$ in this framework leads to a linear system of equations $\mathbf{A} \mathbf{x} = \mathbf{y}$ for the unknown potential values $x_i = \phi_i$, where:
\begin{equation}
a_{ij} = \sum_{T_l \in \mathcal{K}_i} |T_l| \, \nabla \! B_{il}^\top \nabla \! B_{jl} \enspace, \quad y_i = \sum_{T_l \in \mathcal{K}_i} |T_l| \, \nabla \! B_{il}^\top \, \mathrm{v}_l \enspace.
\end{equation}
This system is sparse since the sum for coefficients $a_{ij}$ is non-null iff $\{i, j\} \in \mathcal{K}$ (it is an edge). The sum is then simply over the triangles $T_l$ (two if not a boundary edge, one otherwise) sharing this edge.

This equation can be interpreted as seeking for a potential field $\phi$ whose gradient $\nabla \! \phi$ matches the guide vector field $V$. If $V$ is conservative, i.e., it is the gradient of an existing potential field $\phi$, then $\phi$ is the exact solution. Otherwise, a more general minimizer can still be obtained by least squares but its gradient differs from $V$. In addition, we regularize the minimization via screening:
\begin{equation}
\min_{\mathbf{x} \in \mathbb{R}^N} {\left\Vert \mathbf{A} \mathbf{x} - \mathbf{y} \right\Vert}_2^2 + \lambda {\left\Vert \mathbf{x} - \mathbf{x}' \right\Vert}_2^2 \enspace,
\end{equation}
where $\lambda > 0$ and $\mathbf{x}'$ defines a guide potential field $\phi'$ as $\phi'_i = x'_i$.

\subsection{Texture blending}

\begin{figure}[t]
\centering
\begin{tabular}{@{}c@{}c@{}}
\includegraphics[height=5.6cm]{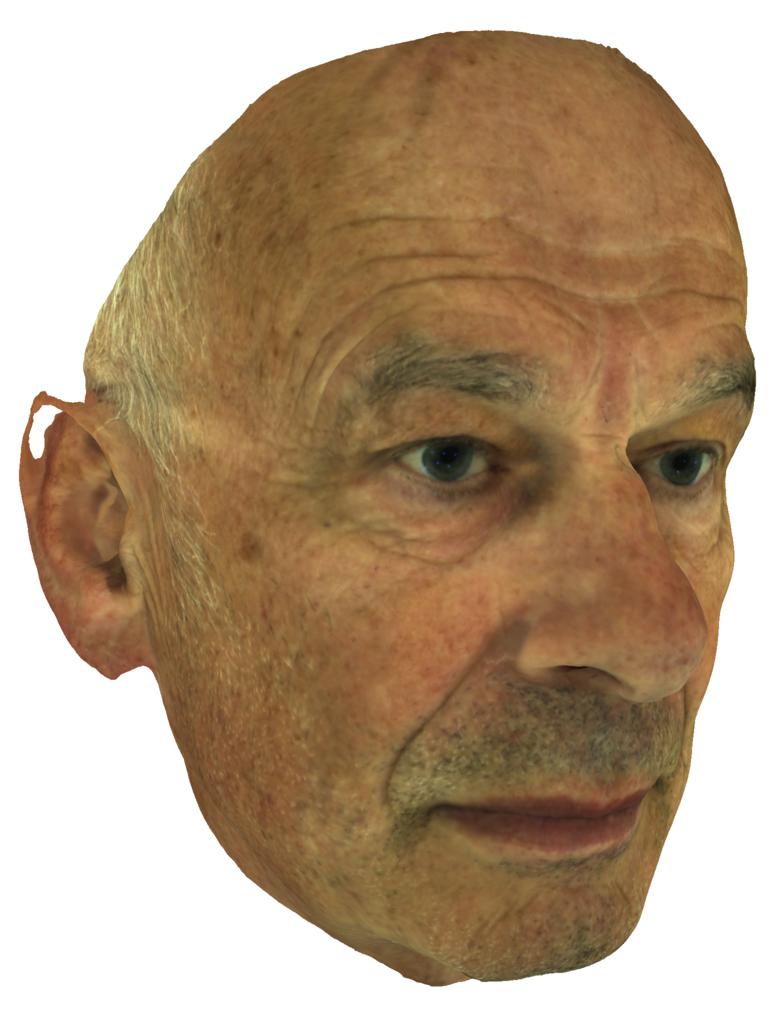} &
\includegraphics[height=5.6cm]{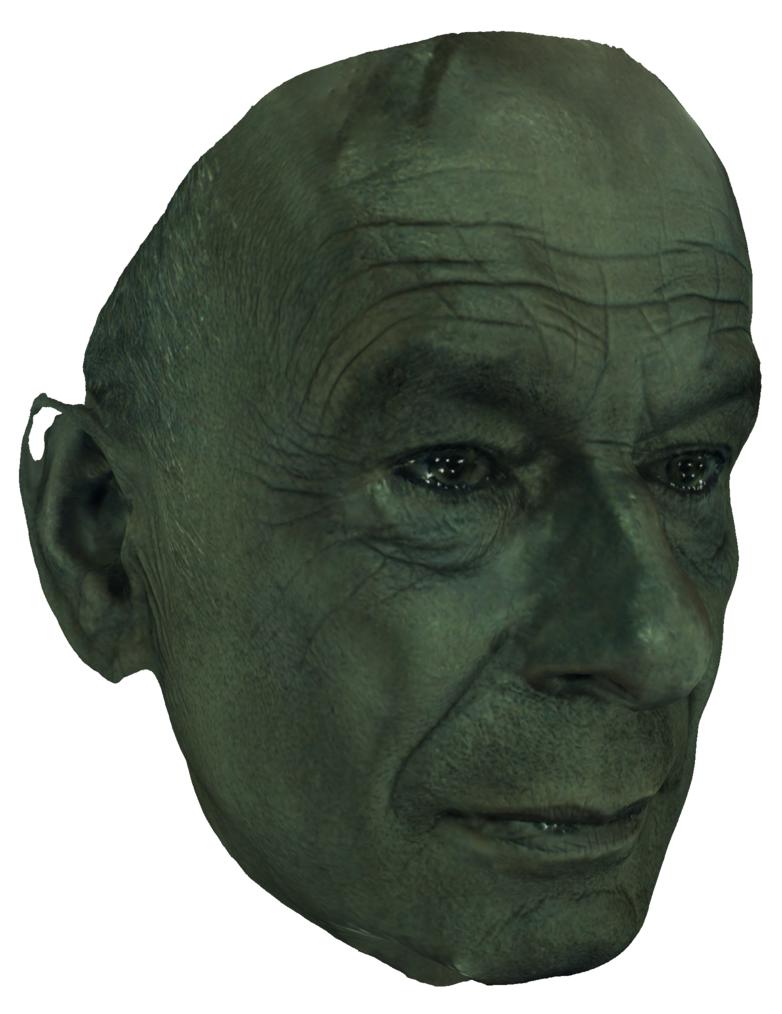}  \\
(a) & (b) \\
\end{tabular}
\caption{Results of our patch-based gradient stitching for (a) diffuse and (b) specular albedo.}
\label{fig:texturing}
\end{figure}

We apply this to solve for texture by considering each color channel independently as a potential field $\phi$. For each view $v$, we compute the mean viewing angle of vertices in the different patches. Unobserved vertices, due either to occlusion or missing information, are assumed to have a viewing angle of $\pi/2$. Hence, patches with unobserved data are penalized and no difference on the nature of non-observability is made. For each patch now, we select texture from the view where the patch has the smallest viewing angle. For unobserved vertices, we also select texture from subsequent sorted views. We end up with partial textures $\phi^{(v)}$ that we stitch in overlaps by Poisson blending. To build up the guide vector field $V$, we select local texture gradients by least angle for each triangle $T_l$:
\begin{equation}
\mathrm{v}_l = \sum_{i \in \mathcal{K}} \phi_i^{(v_l)} \, \nabla \! B_{il}\enspace,
\end{equation}
where $v_l$ is the view whose angle is minimal for triangle $T_l$. We also fill in unobserved faces simply by setting their gradients to zero for smoothness. Screening is done via a rough estimate $\phi'$ obtained by averaging textures $\phi^{(v)}$, unobserved textures being discarded from the regularization. We use a small penalty $\lambda \!\! = \!\! 10^{-6}$ to remove color offset indeterminacies since we did not observe dramatic color bleeding issues compared to~\cite{Chuang2009}. We show in Figure~\ref{fig:texturing} the results of the gradient stitching on the diffuse and specular textures.

\subsection{Surface normal blending}

Ultimately, our goal is to transfer the detail from the photometric normal maps to the mesh surface. One approach to this problem would be to start by stitching the normal maps from each view into a seamless and complete normal map for the whole face using the texture stitching approach above. Then, the normals could be embossed onto the mesh using an algorithm such as Nehab's \cite{Nehab:05}. There are two drawbacks to this approach. First, since normal maps are fields of unit vectors, the stitching must preserve unit length. Hence, the linear least squares solution used for textures would need to include quadratic equality constraints. This amounts to a quadratically constrained quadratic program which is no longer a convex optimisation problem. Second, the stitched texture will not necessarily correspond to a real surface. That is to say, the normals would not satisfy an integrability constraint.

We solve both of these problems by proposing a method to simultaneously stitch the normals and transfer the detail to the mesh. We do so using the same patch-based approach as for texture data and hence provide a unifying framework for Poisson blending both texture and shape using patches.

Instead of stitching in the surface normal domain, we solve for the mesh whose surface normals best fit those selected from the photometric normal maps by the patch-based selection approach. Our guidance field takes the form of per-vertex surface normals.
We begin by writing the Laplace-Beltrami operator as applied to the mesh coordinate function at a vertex ${\bf v}_i$ and note the relationship to the surface normal direction:
\begin{equation}
 \Delta({\bf v}_i)=\frac{1}{2|\Omega_i|}\sum_{\left\{j|\{i,j\}\in\cal{K}\right\}} w_{ij}({\bf v}_i-{\bf v}_j) = -H({\bf v}_i){\bf n}_i
 \end{equation}
where the weights are $w_{ij}=\cot\alpha_{ij}+\cot\beta_{ij}$, $\Omega_i$ is the area of the Voronoi cell of $i$ and $\alpha_{ij}$ and $\beta_{ij}$ are the two angles opposite the edge $\{i,j\}$.

We cannot directly apply mesh editing techniques to our problem. If the mean curvature normal was known at each vertex, our problem would simply be a Laplacian mesh editing problem.
Instead, we know only the unit surface normal. However, we propose a linearisation inspired by the direct linear transformation (DLT) algorithm \cite{Hartley:03}.
We can obtain a linear system of equations by noting that the Laplacian coordinates and surface normal differ only by a scale factor:
\begin{equation}
 \sum_{\left\{j|\{i,j\}\in\cal{K}\right\}} w_{ij}({\bf v}_i-{\bf v}_j) \sim {\bf n}_i
\end{equation}
where $\sim$ denotes equality up to a non-zero scalar multiplication. Such sets of relations can be solved using the DLT. The idea is to minimise the cross product between the differential coordinates and the target normals. This has the nice property of giving higher weight to regions of high curvature so our method will seek to preserve high frequency detail.
Accordingly, we write
\begin{equation}
\left[
{\bf n}_i
\right]_{\times}
\sum_{\left\{j|\{i,j\}\in\cal{K}\right\}} w_{ij}({\bf v}_i-{\bf v}_j)={\bf 0},
\end{equation}
where ${\bf 0}=[0\ 0\ 0]^T$ and $\left[ . \right]_{\times}$ is the cross product matrix:
\begin{equation}
\left[ {\bf x} \right]_{\times} = \left[ \begin{array}{ccc}
0 & -x_3 & x_2 \\
x_3 & 0 & -x_1 \\
-x_2 & x_1 & 0 \\
\end{array}
\right].
\end{equation}
Hence, each vertex normal contributes three linear equations, leading to a large, sparse system of linear equations. The solution is ambiguous up to a scale factor and we wish to retain the low frequency characteristics of the base shape. Hence, as for texture, we add a screening term that penalises departure from the vertex positions of the base mesh. In practice, we give the screening term a very low weight to maximise detail transfer from the normal maps.

\begin{figure}[t]
\centering
\begin{tabular}{cc}
\includegraphics[height=3cm]{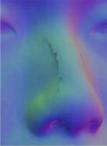} &
\includegraphics[height=3cm]{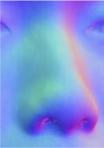}  \\
(a) & (b) \\
\end{tabular}
\caption{Normal stitching before (a) and after (b) Poisson blending}
\label{fig:normal_blending}
\end{figure}

In Figure \ref{fig:normal_blending} we compare Poisson blending of the surface normals with naive back projection of the least angle patch. Without the blending, the seams are clearly visible at patch boundaries when the view from which they are selected changes. After Poisson blending, we show the normals of the refined mesh where it is clear the transition between views is smooth.

\subsection{View Dependant Fresnel Gain}\label{fresnel_gain}

The measured specular albedo has a view dependency which makes it unreliable when the viewing angle is large, particularly when it is close to glancing angles. This is because the proportion of light that is specularly reflected is dependent on the angle of incidence following Fresnel's equations. The effect is that specular albedo appears amplified towards the occluding boundary, so called ``Fresnel gain''.

In applications where the whole specular albedo is to be used, it is important to correct these Fresnel effect to achieve multi-view photometric consistency. Previous work has either simply cropped the face to exclude regions of unreliable specular reflectance \cite{stratou2011effect}
or attempted a data-driven correction process.
 Ghosh \etal \cite{ghosh2011multiview} took this latter approach. They bin specular albedo values into a histogram as a function of viewing angle and fit a smooth function to the measured data. Specular albedo is then scaled at each vertex down to the average gain at zero-view angle. The problem with this approach is that it makes the assumption that all specular parameters, including specular albedo, are constant over the face surface. Although the approach succeeds in removing extreme values close to the boundary, its accuracy is questionable and the resulting specular albedo is unlikely to be an accurate  measurement of an intrinsic property.

We take an alternative pragmatic approach which in practice yields seamless specular albedo maps without the lack of physical motivation or fragility of applying a correction function.
Since our patch selection is based on least viewing angle, the patches selected by our stitching method typically have average viewing angle $<\frac{\pi}{2}$ radians for all patches. At these viewing angles, Fresnel gain is negligible and we therefore prevent selecting patches containing regions at grazing angles. This is an advantage of using multiple photometric views: we ensure that all regions of the face are observed at small viewing angles in at least one image. A stitched specular albedo map is shown in Figure \ref{fig:texturing}(b). Artefacts due to Fresnel gain at the boundary are successfully removed. 

\section{Experimental Results}

We now present results of applying our face capture pipeline to a set of faces of varying age, ethnicity and gender. Our results are obtained using a custom built light stage comprising 41 ultra bright white LEDs mounted on a geodesic dome of diameter 1.8m. The photometric camera is a Nikon D200 in front of which we mount an LC-Tec FPM-L-AR optoelectric polarising filter. Each LED has a rotatable linear polarising filter in front of it. Their orientation is tuned by placing a sphere of low diffuse albedo and high specular albedo (a black snooker ball) in the centre of the dome and adjusting the filter orientation until the specular reflection is completely cancelled in the camera's view. LED brightness is controlled via PWM from an MBED micro controller which also controls camera shutters and the polarisation state of the photometric camera. The multiview cameras are Canon 7Ds. A complete capture sequence takes around 3 seconds and this is repeated for three poses.

\begin{figure}[t]
\centering
\setlength{\tabcolsep}{2.3pt}
\begin{tabular}{cc}
Specular Normals & Rendering \\
\includegraphics[width=3cm]{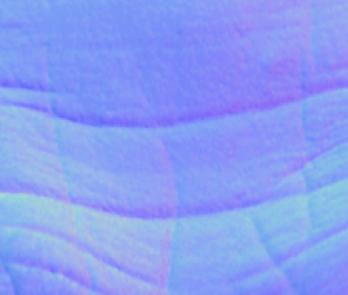} &
\includegraphics[width=3cm]{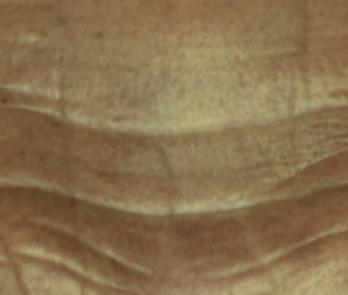} \\
\includegraphics[width=3cm]{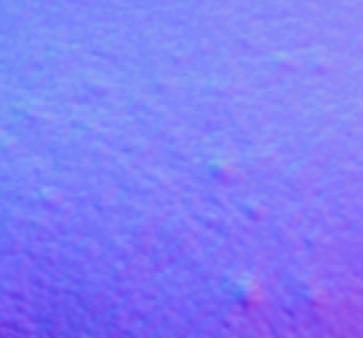} &
\includegraphics[width=3cm]{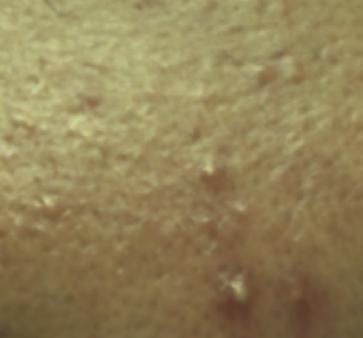} \\
\includegraphics[width=3cm]{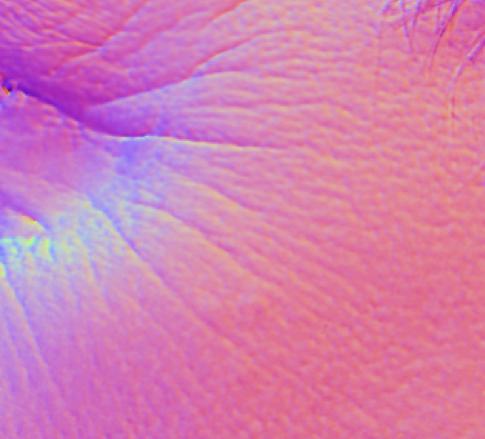} &
\includegraphics[width=3cm]{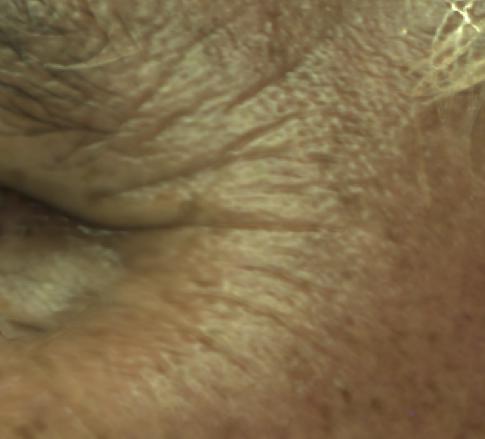} \\
\includegraphics[width=3cm]{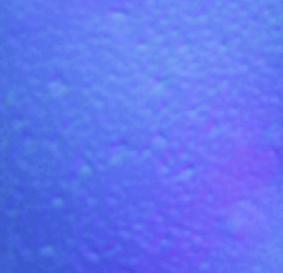} &
\includegraphics[width=3cm]{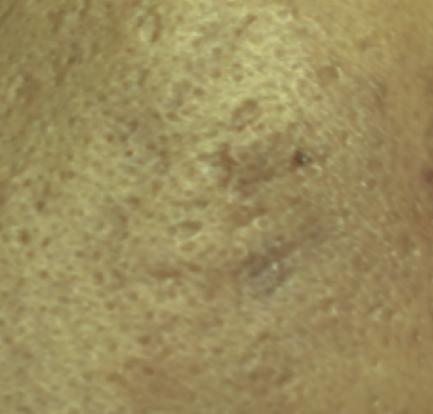} \\
\end{tabular}
\setlength{\tabcolsep}{6pt}
\caption{Detail renderings of different facial regions.}
\label{fig:rendering_regions}
\end{figure}

\subsection{Intrinsic texture stitching}

\begin{figure*}
\centering
\begin{tabular}{ccc}
 {\footnotesize No Blending, No Overlap} & {\footnotesize No Blending, with Overlap} & {\footnotesize Overlap and Blending} \\
\includegraphics[height=5cm]{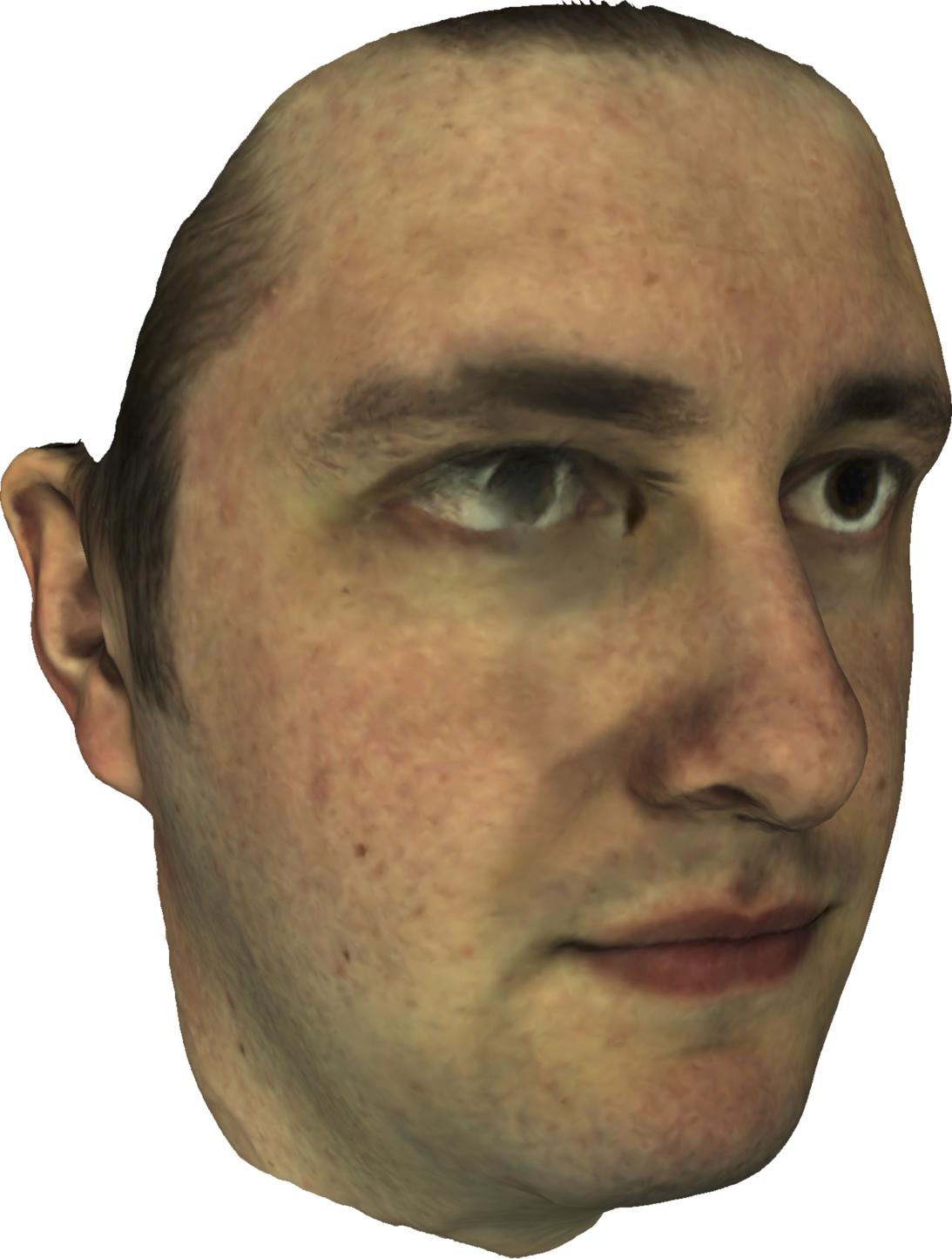} &
\includegraphics[height=5cm]{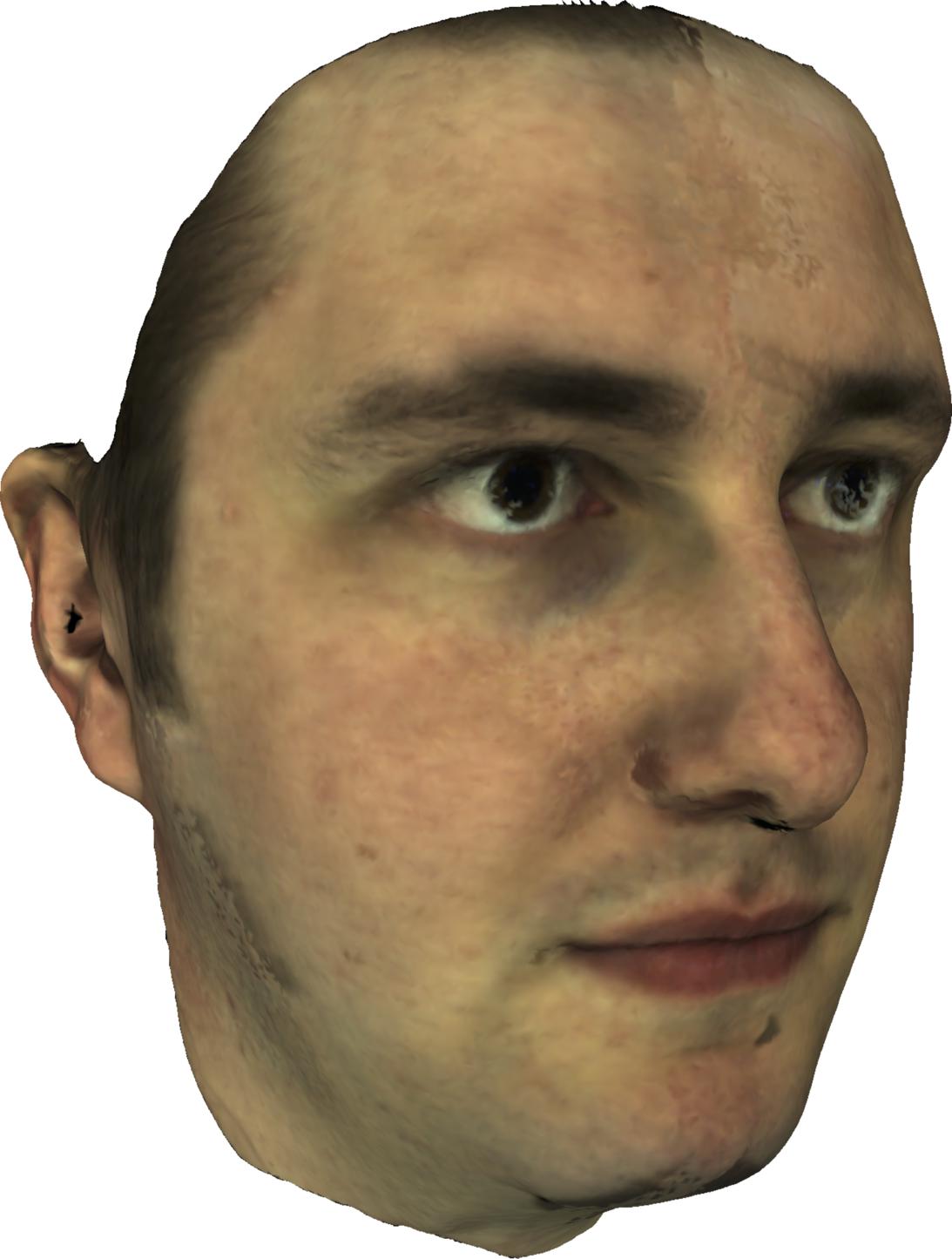} &
\includegraphics[height=5cm]{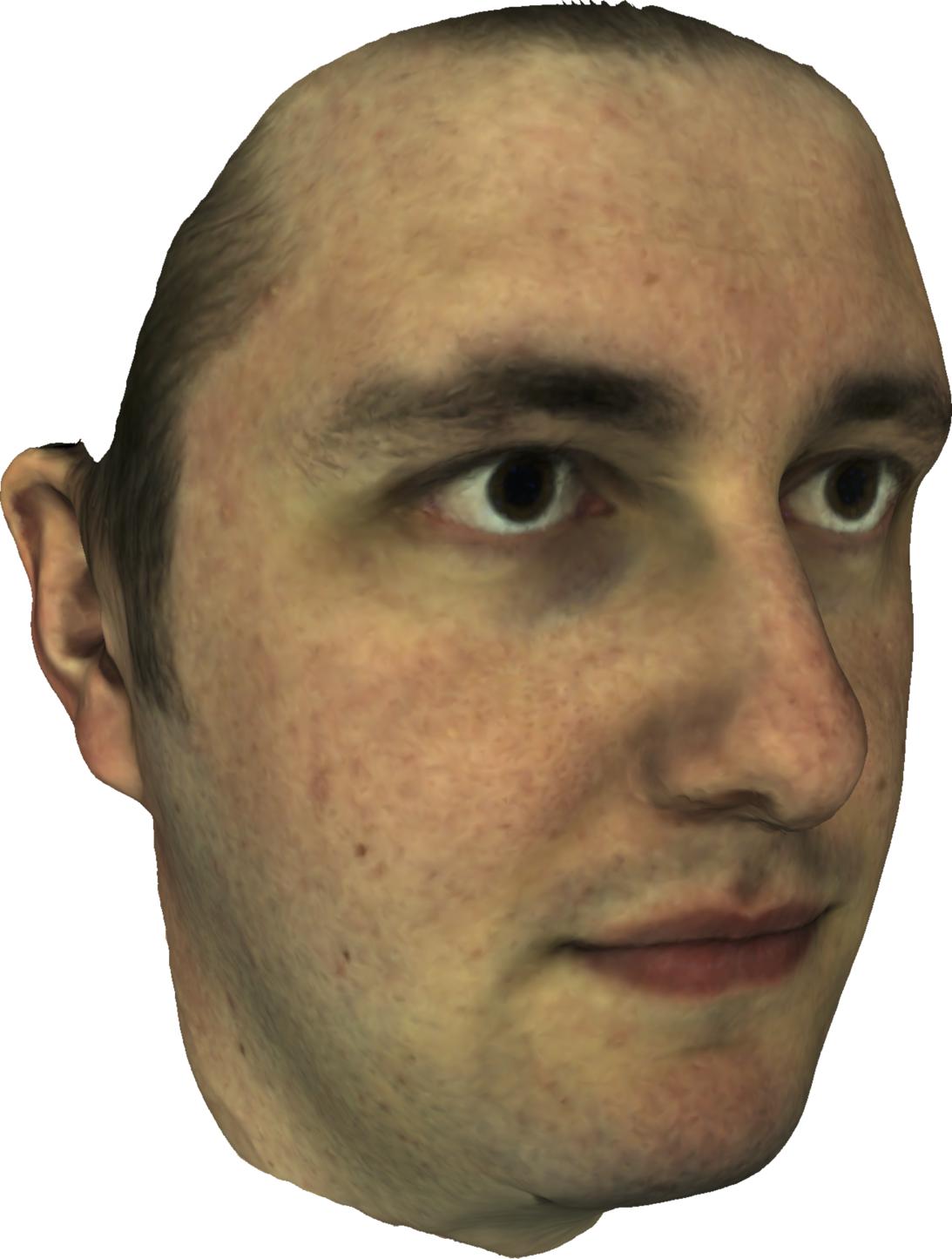} \\
\end{tabular}
\caption{Patch-based texture stitching with different configurations.}
\label{fig:blending}
\end{figure*}

In Figure \ref{fig:blending} we show the results of our patch-based
texture stitching process. In this case, we show the
results of stitching diffuse albedo maps from the three
photometric views. In all results we use a segmentation consisting of 100 patches. On the left, we show a result
where non-overlapping patches are copied directly
from the best view without blending. In the mid-
dle, we show a result where patches overlap but
no blending is performed (textures are averaged in
the overlapping regions). There are clear artefacts
associated with boundaries between patches and (in
the middle column) loss of detail due to averaging.
On the right, our stitched result contains no patch
boundary artefacts yet retains sharp detail over the
whole face surface.

\begin{figure*}
\centering
\begin{tabular}{ccc}
Base Mesh & Diffuse Normals & Specular Normals \\
\includegraphics[height=5cm]{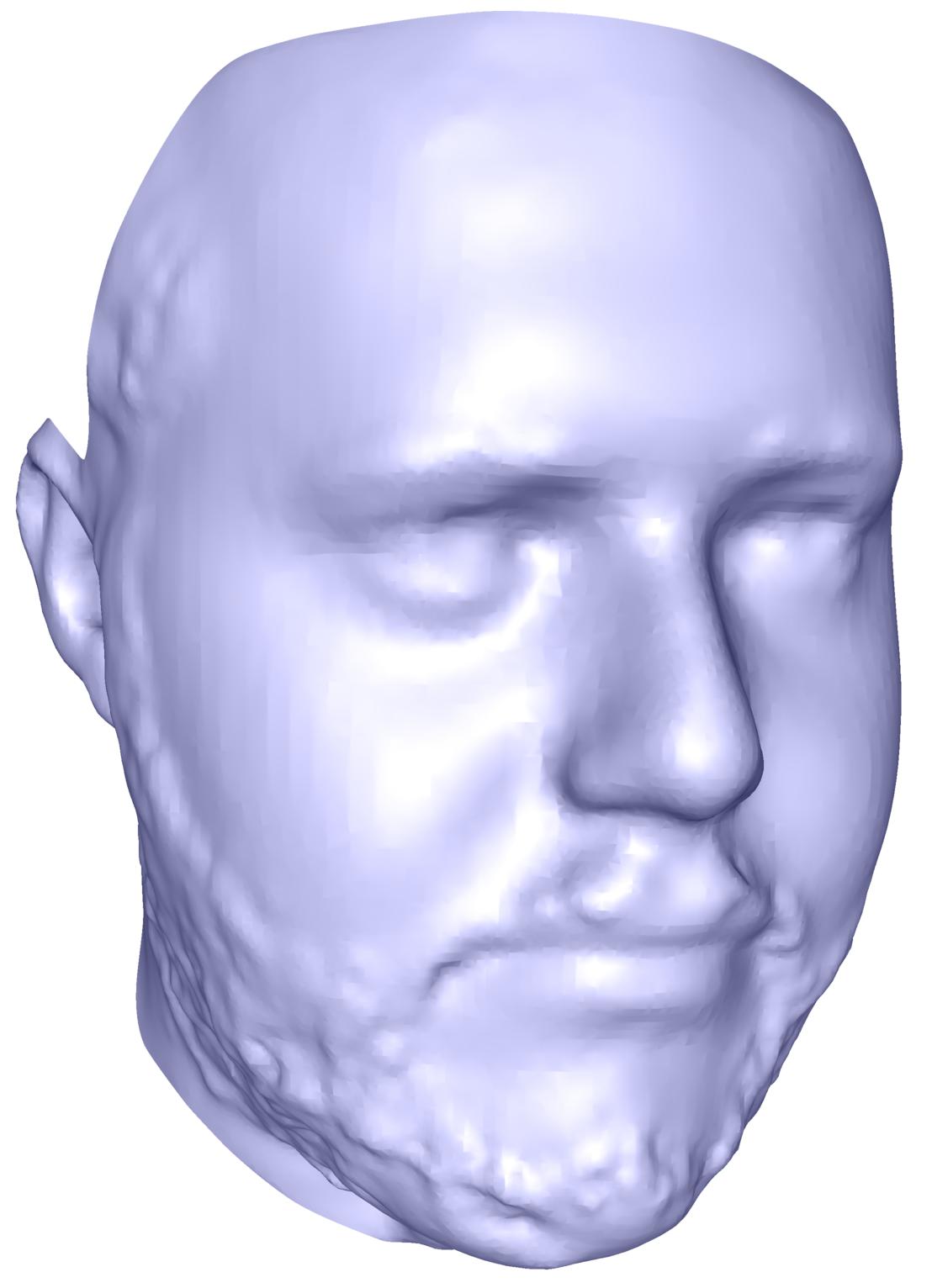} &
\includegraphics[height=5cm]{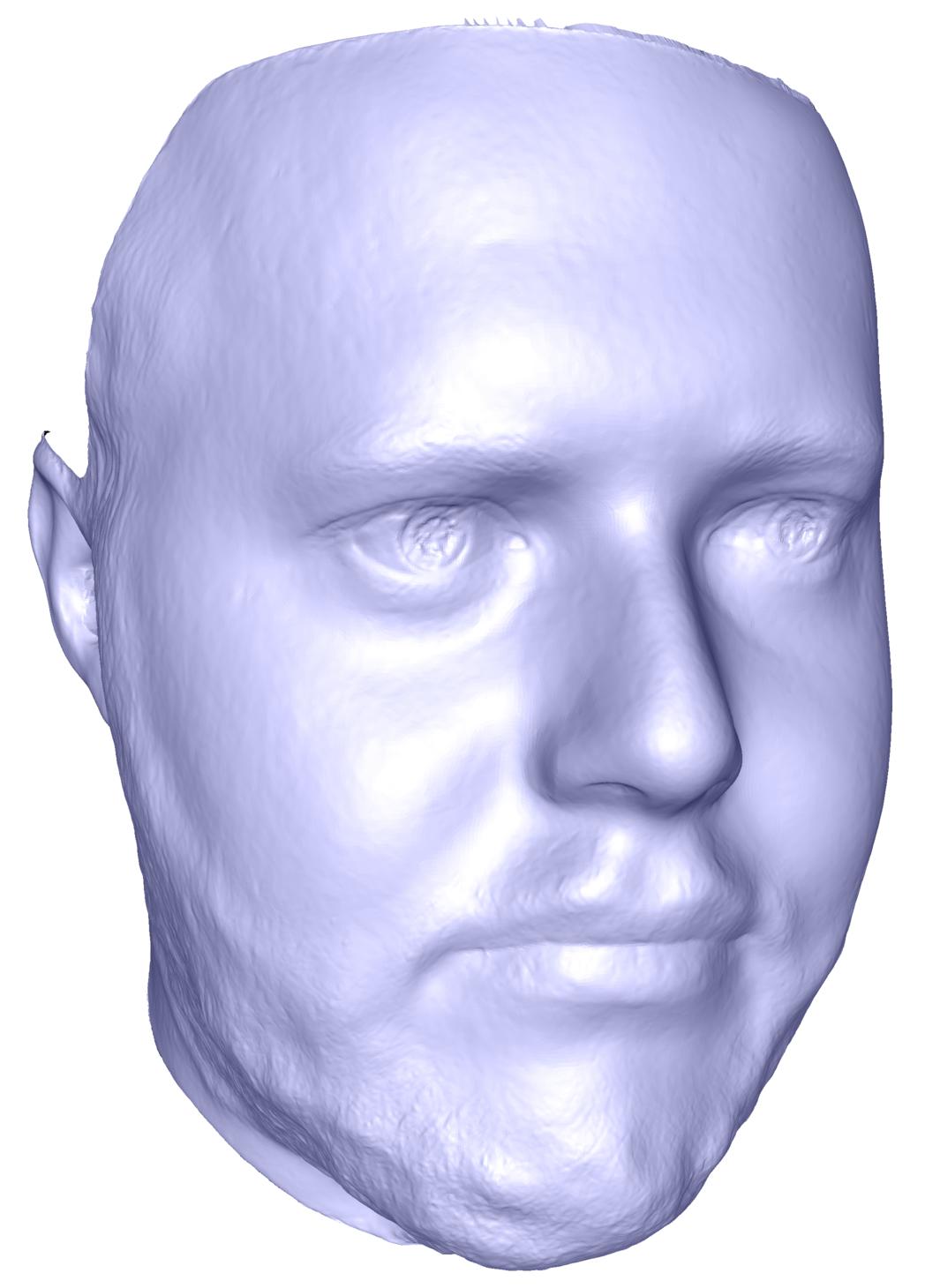} &
\includegraphics[height=5cm]{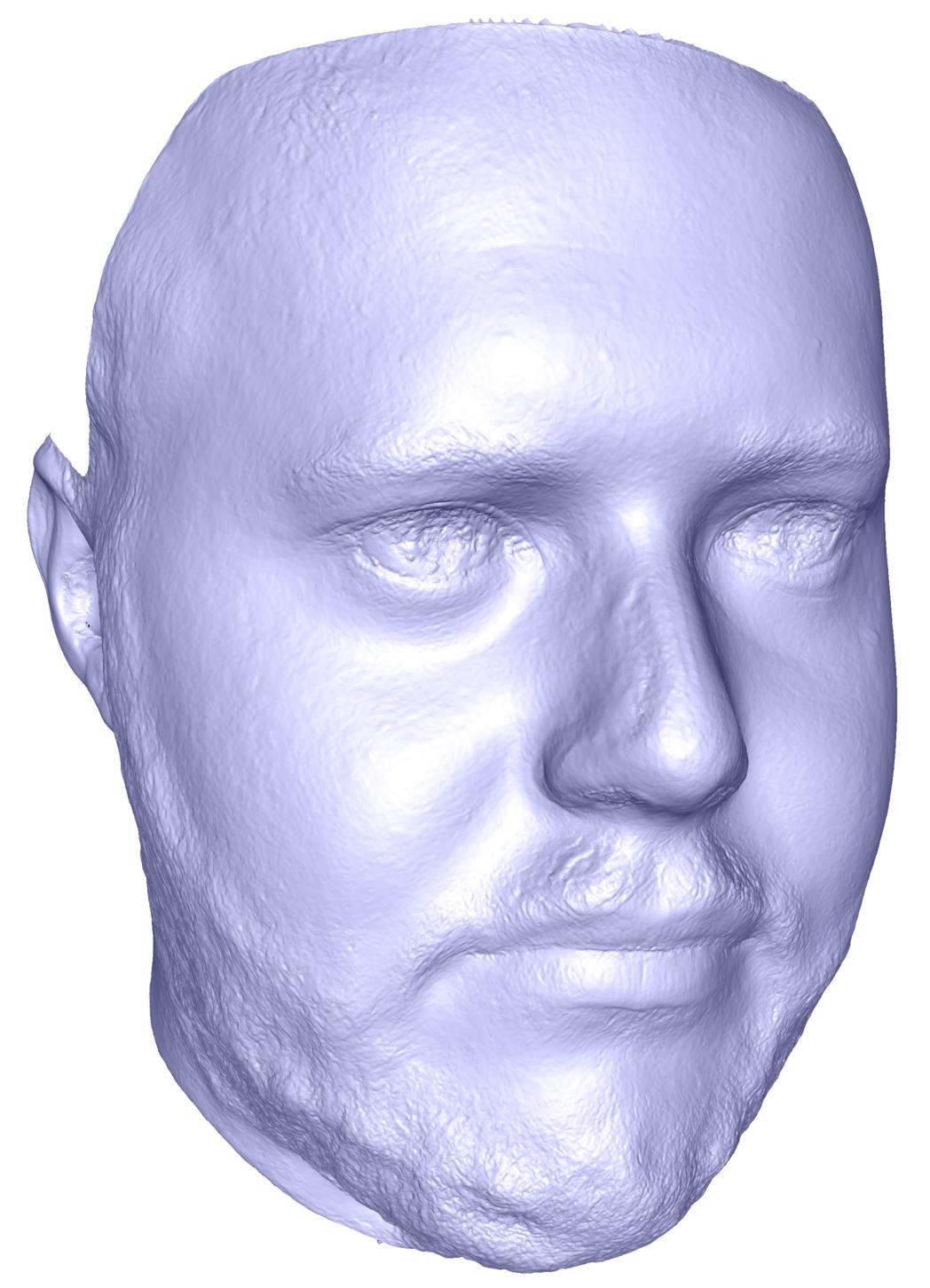} \\
\includegraphics[height=5cm]{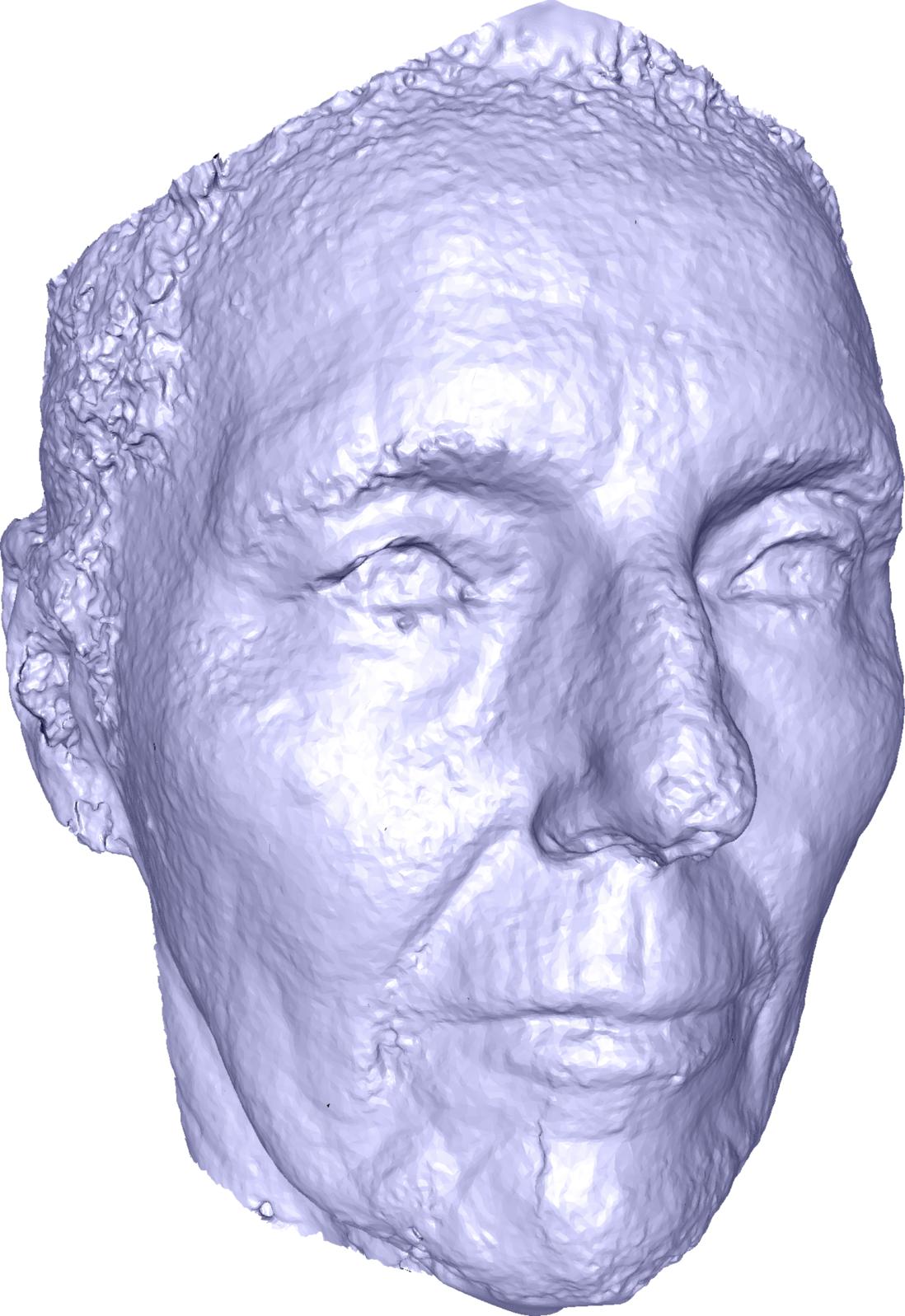} &
\includegraphics[height=5cm]{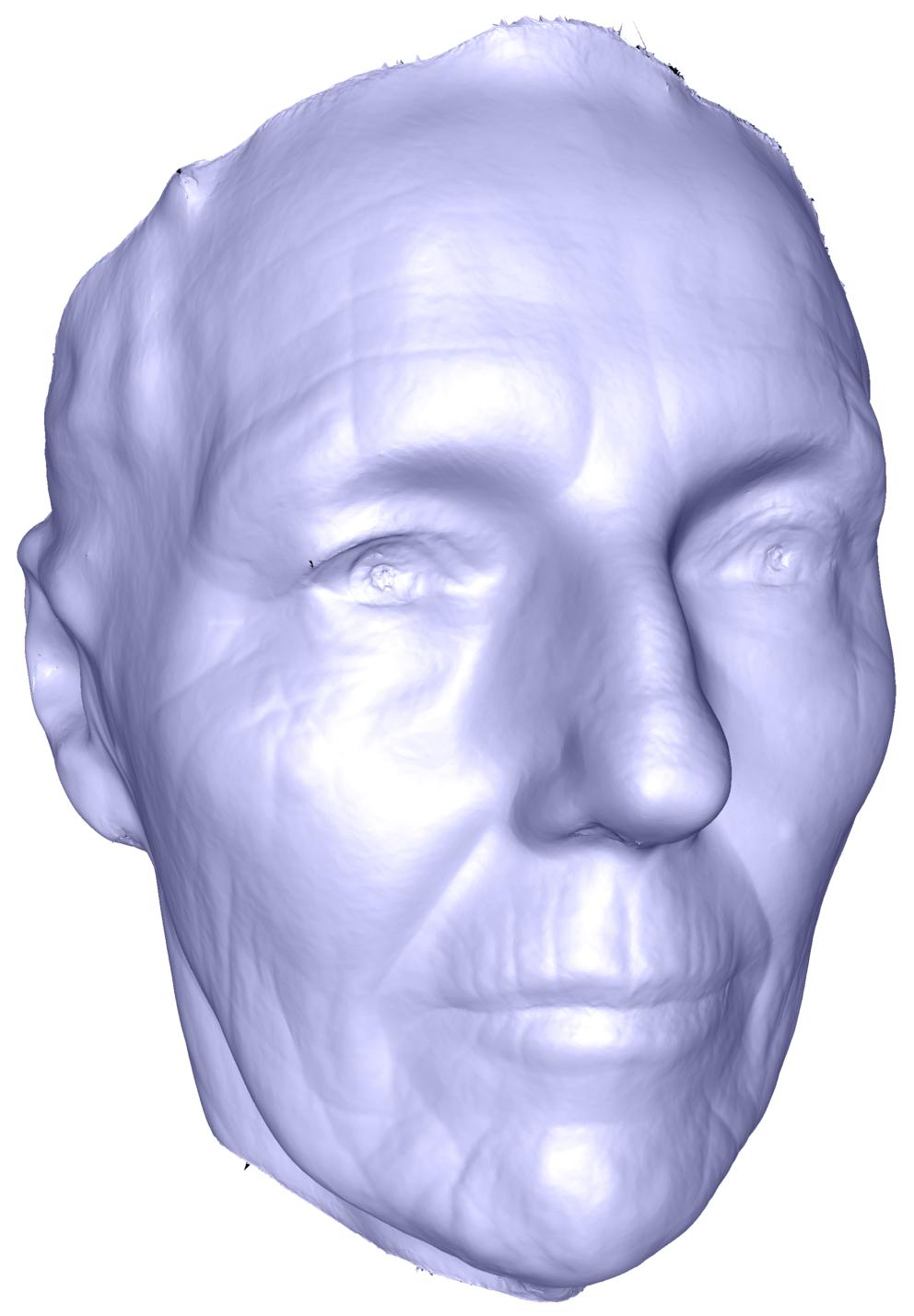} &
\includegraphics[height=5cm]{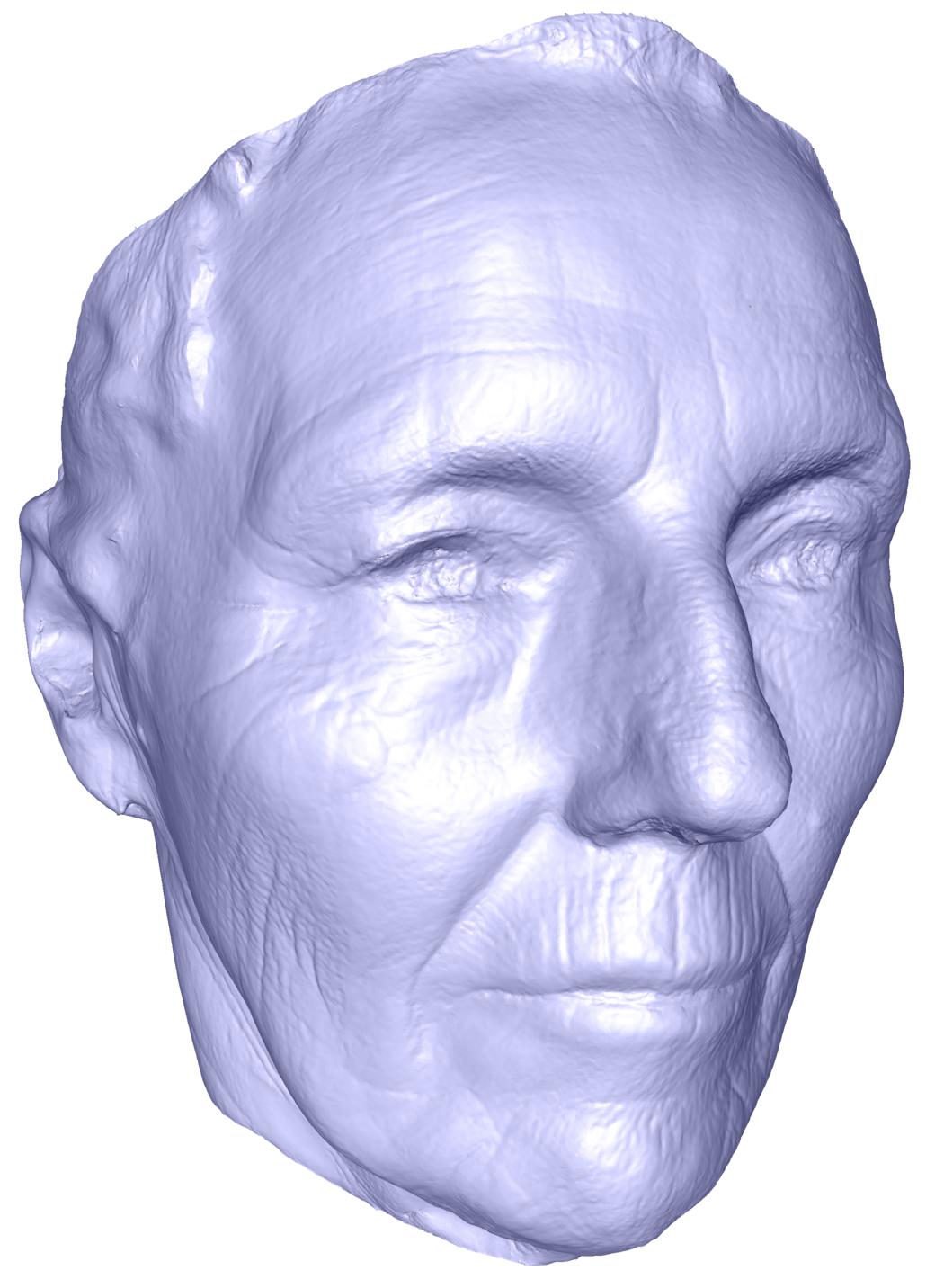} \\
\end{tabular}
\caption{Normal stitching/mesh refinement using diffuse normals (middle column) or specular normals (right column).}
\label{fig:mesh_refinment}
\end{figure*}

\subsection{Normal stitching and detail transfer}

To refine the base mesh either the diffuse or specular
surface normals can be used. Figure \ref{fig:mesh_refinment} shows the results of using each. Note that the base mesh provided
by multiview stereo is coarse and noisy. Once the
mesh has been refined in order to match the stitched
normals, it is evident that the diffuse normal maps
tend to produce smother a mesh while the specular
normal maps yield more surface details. This is consistent with the findings of Ma et al. \cite{ma2007rapid} and is explained
by the fact that the surface reflectance from which
the normals are estimated has different characteristics
depending on whether it is diffuse or specular. In the
diffuse case, the reflected light is considerably affected
by subsurface scattering. We note that our result with
the specular normals achieves a very high level of
detail over the whole face surface.

\subsection{Rendering}

\begin{figure*}

\begin{tabular}{cccc}
\includegraphics[width=4cm]{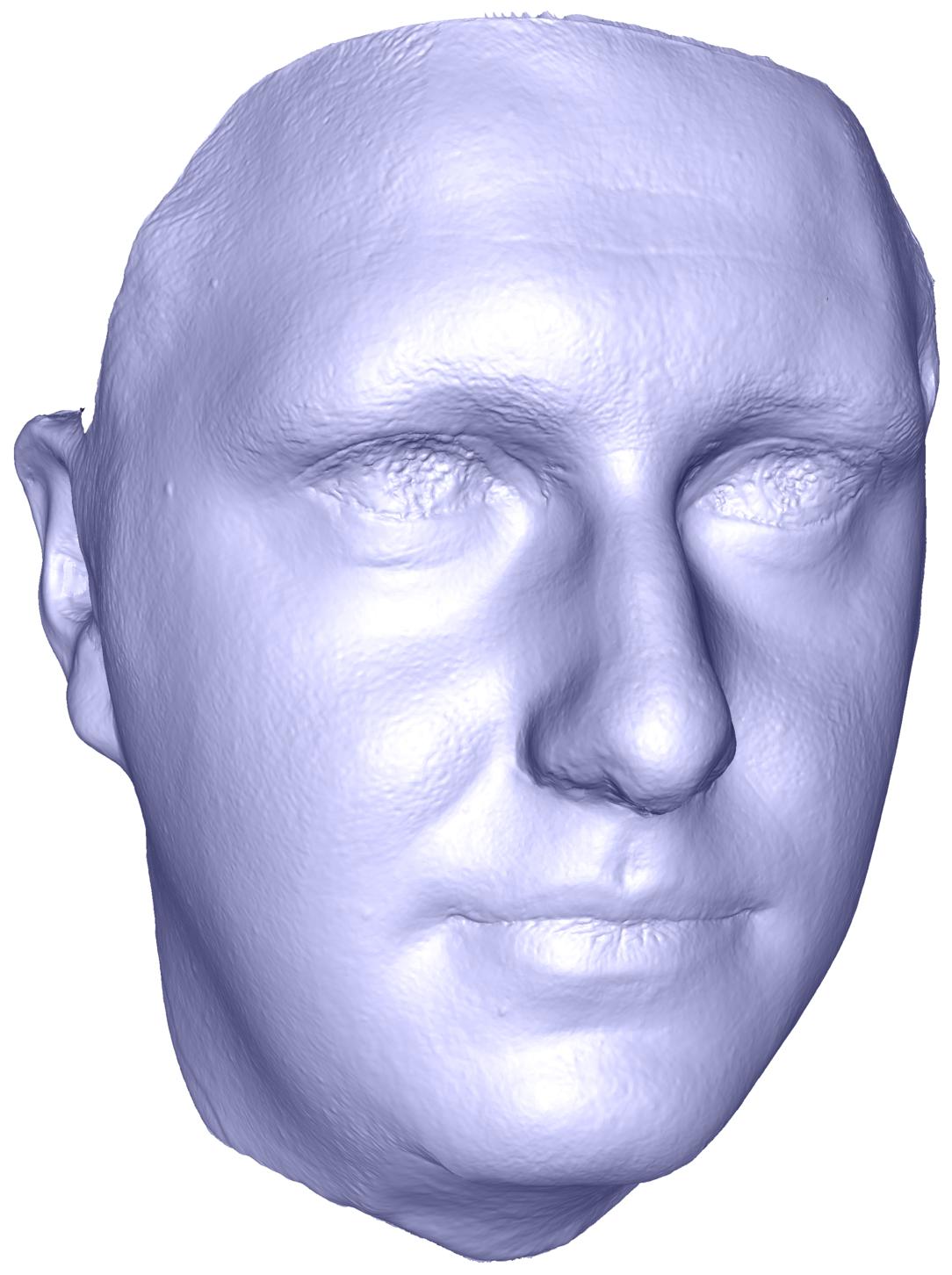} &
\includegraphics[width=4cm]{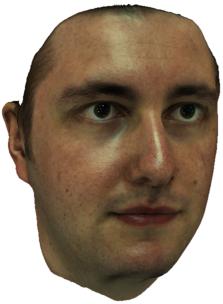} &
\includegraphics[width=4cm]{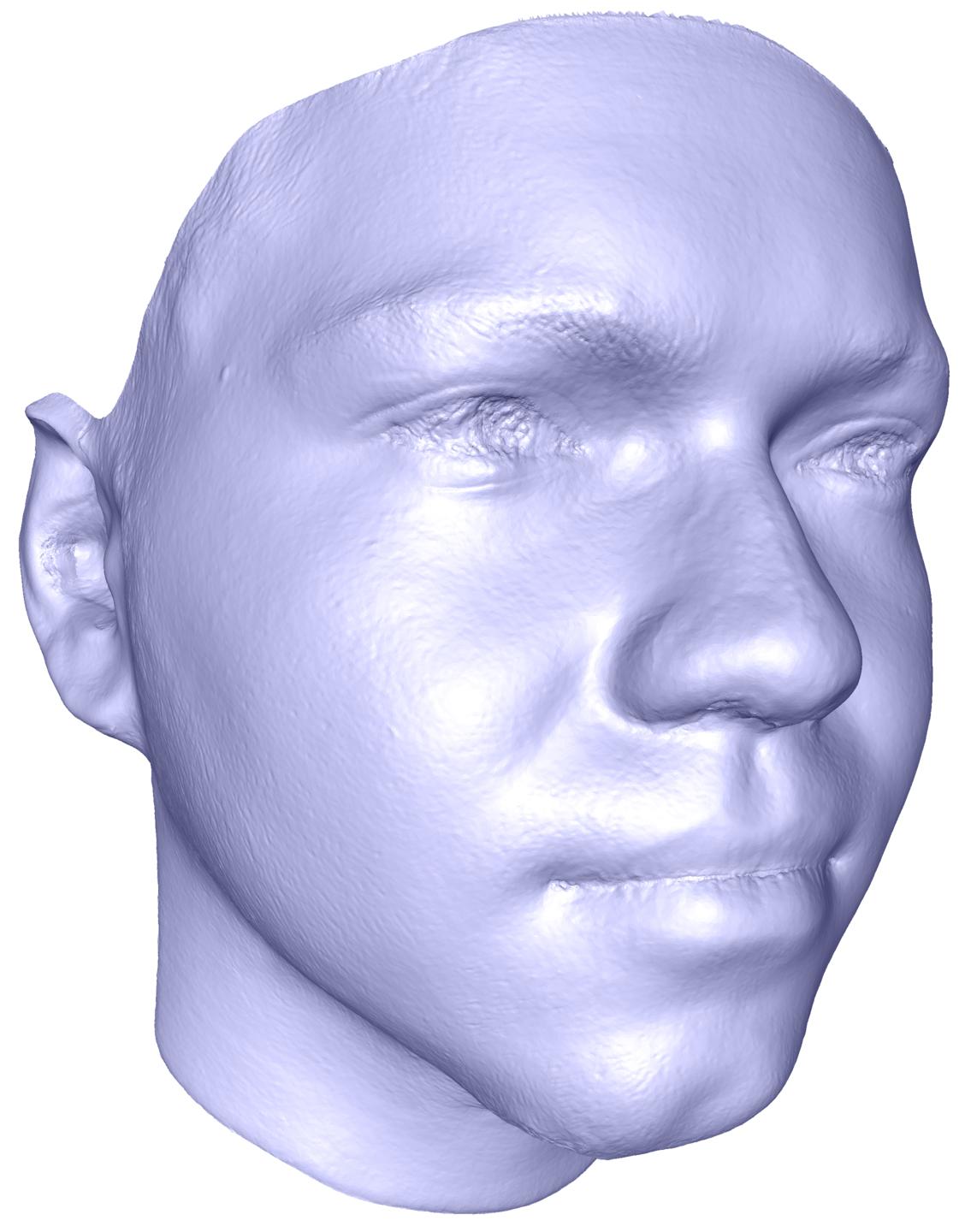} &
\includegraphics[width=4cm]{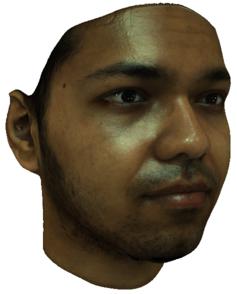}  \\

\includegraphics[width=4cm]{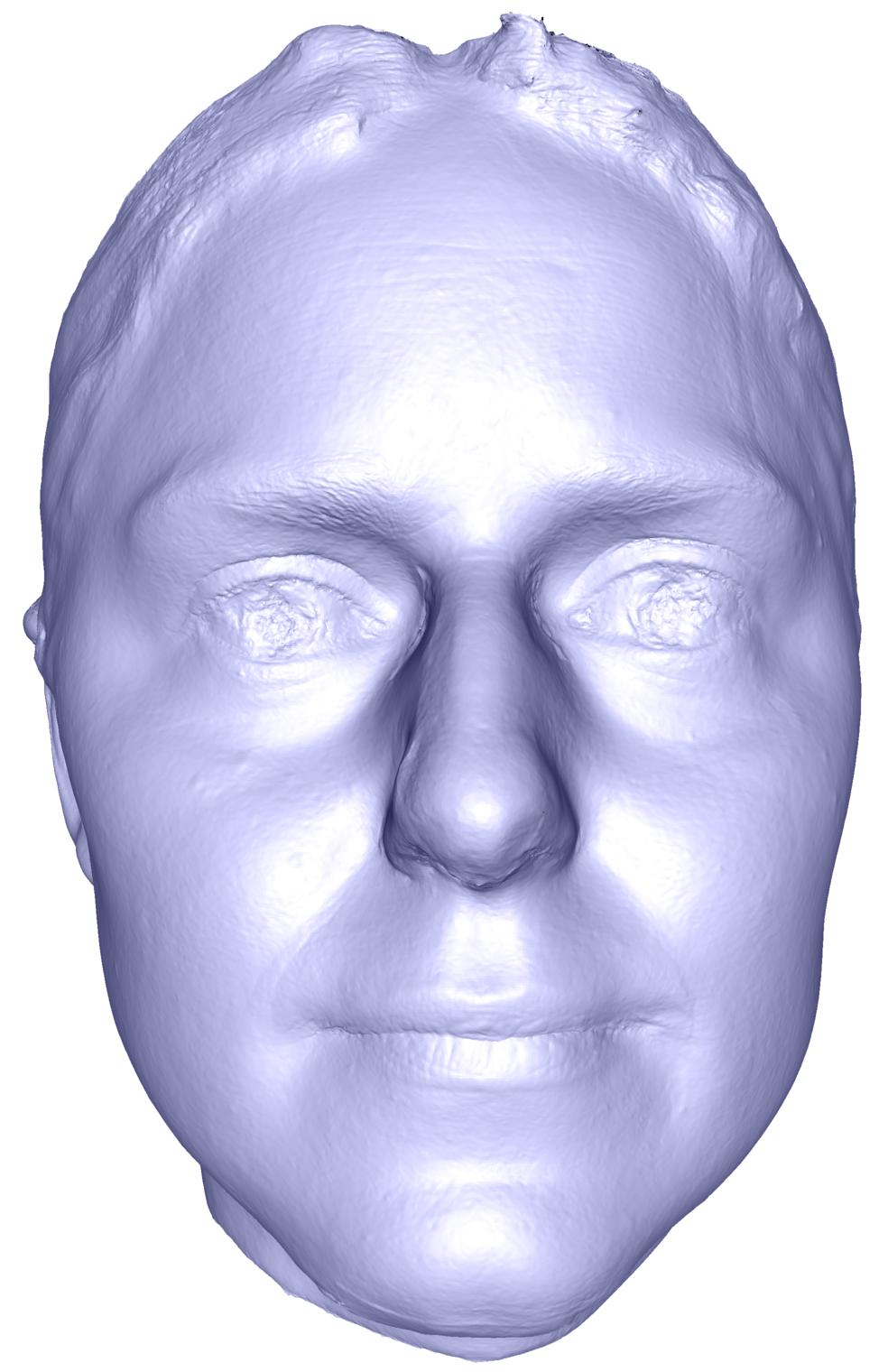} &
\includegraphics[width=4cm]{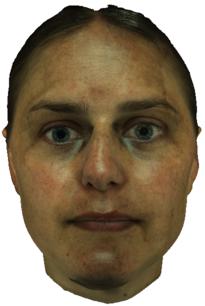} &
\includegraphics[width=4cm]{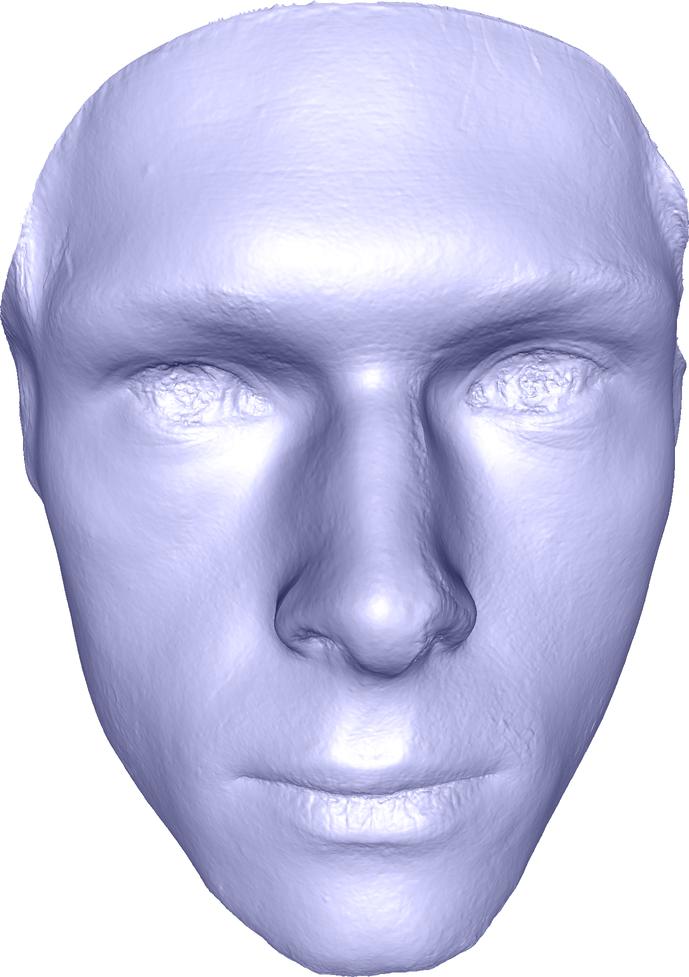} &
\includegraphics[width=4cm]{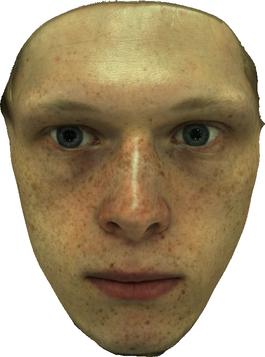}  \\

\includegraphics[width=4cm]{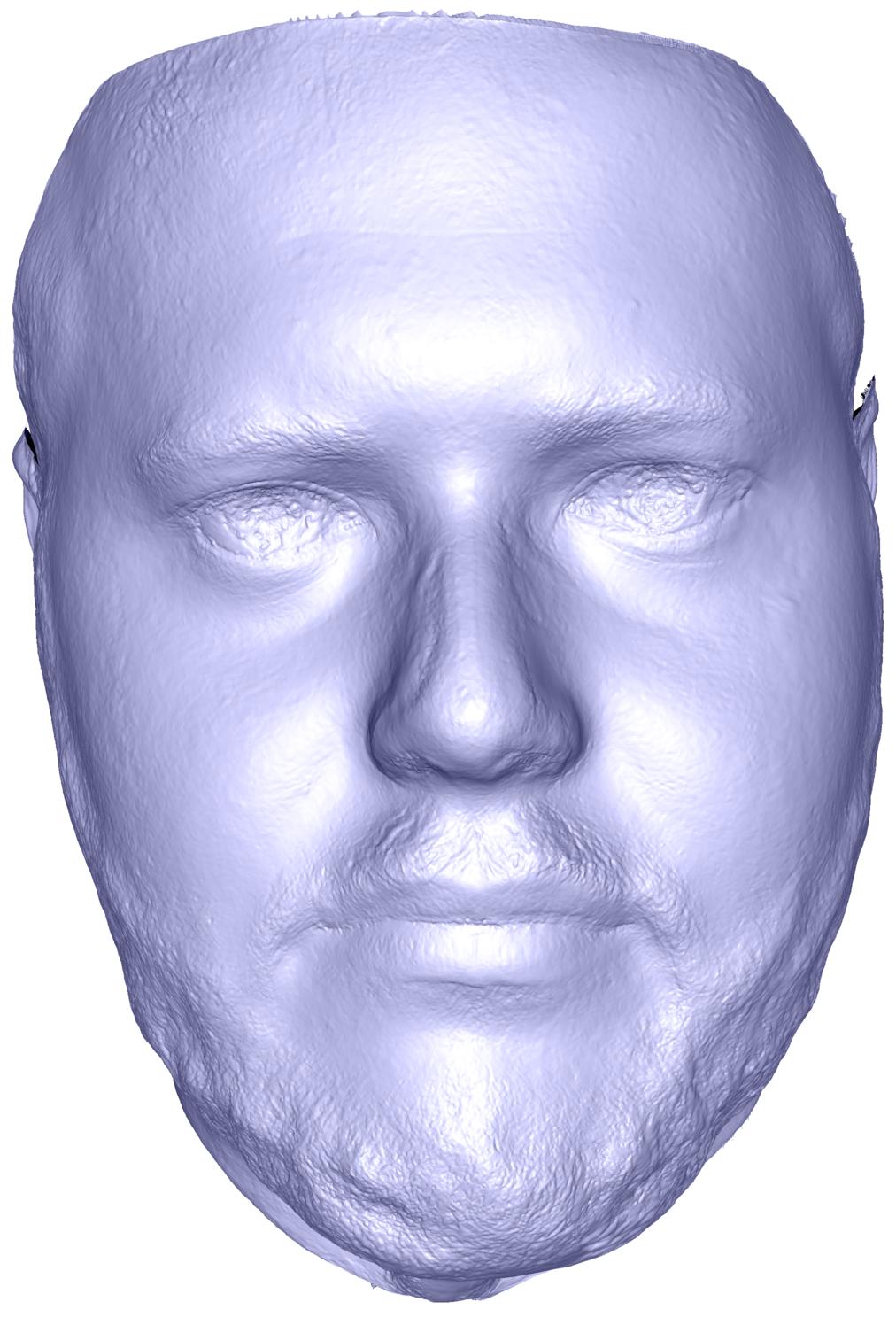} &
\includegraphics[width=4cm]{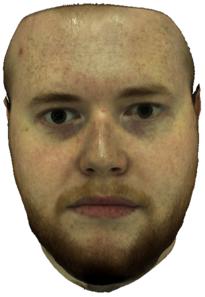} &
\includegraphics[width=4cm]{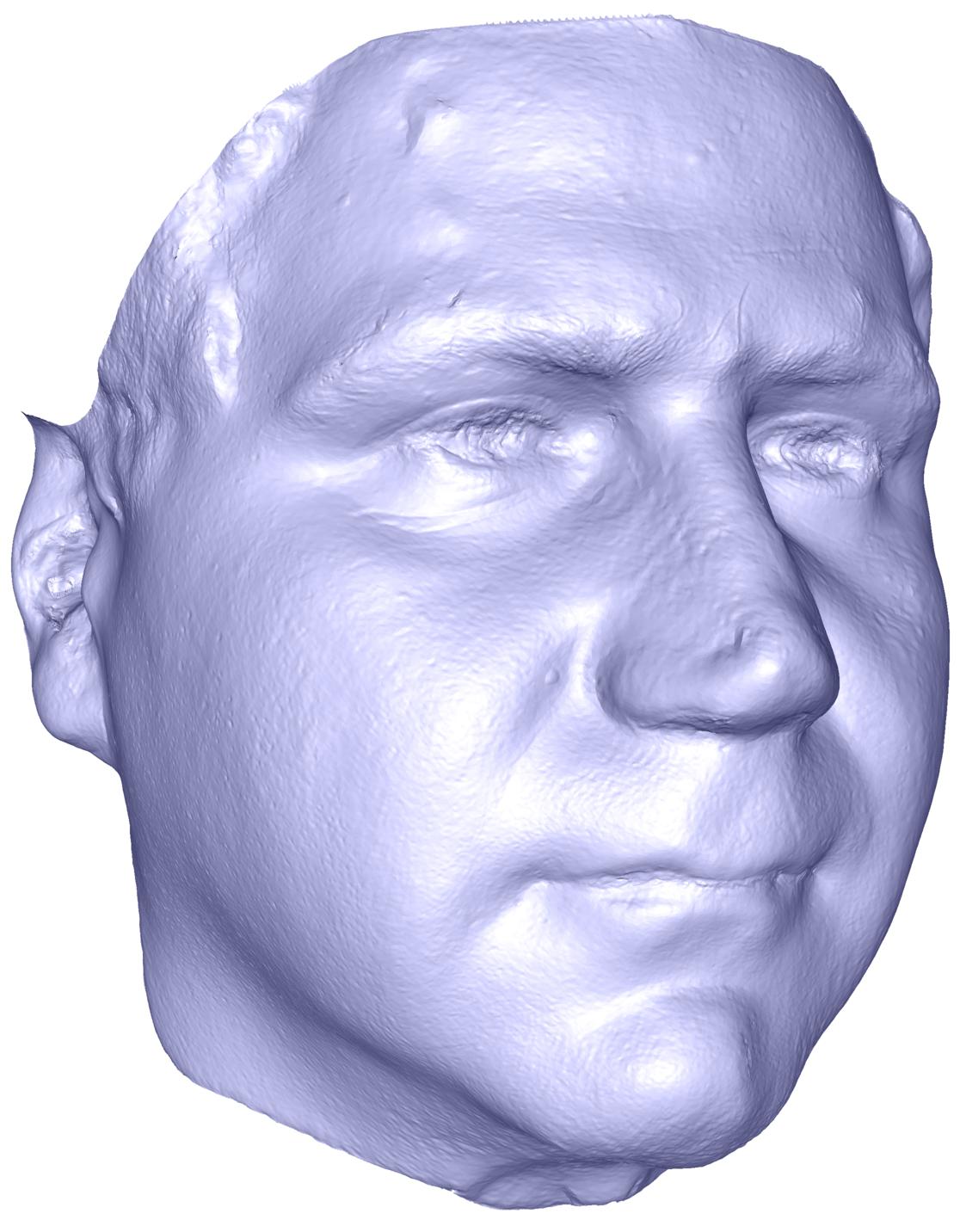} &
\includegraphics[width=4cm]{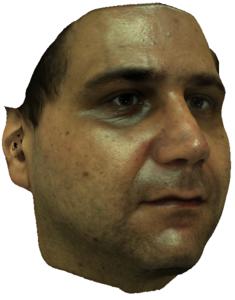}  \\

\includegraphics[width=4cm]{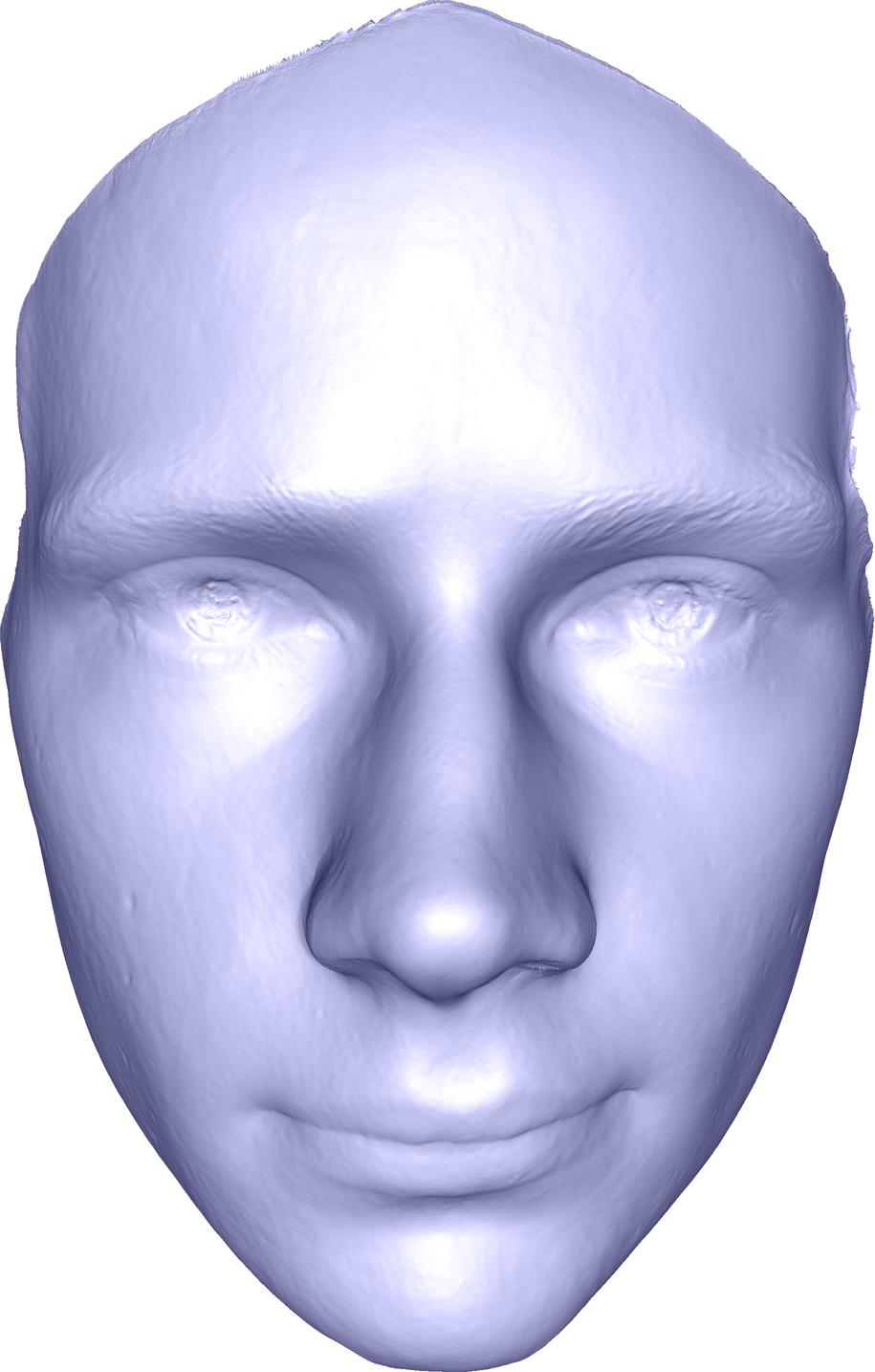} &
\includegraphics[width=4cm]{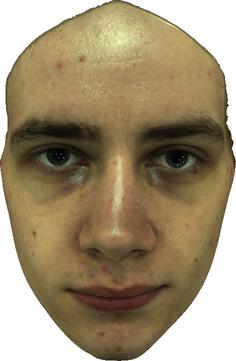} &
\includegraphics[width=4cm]{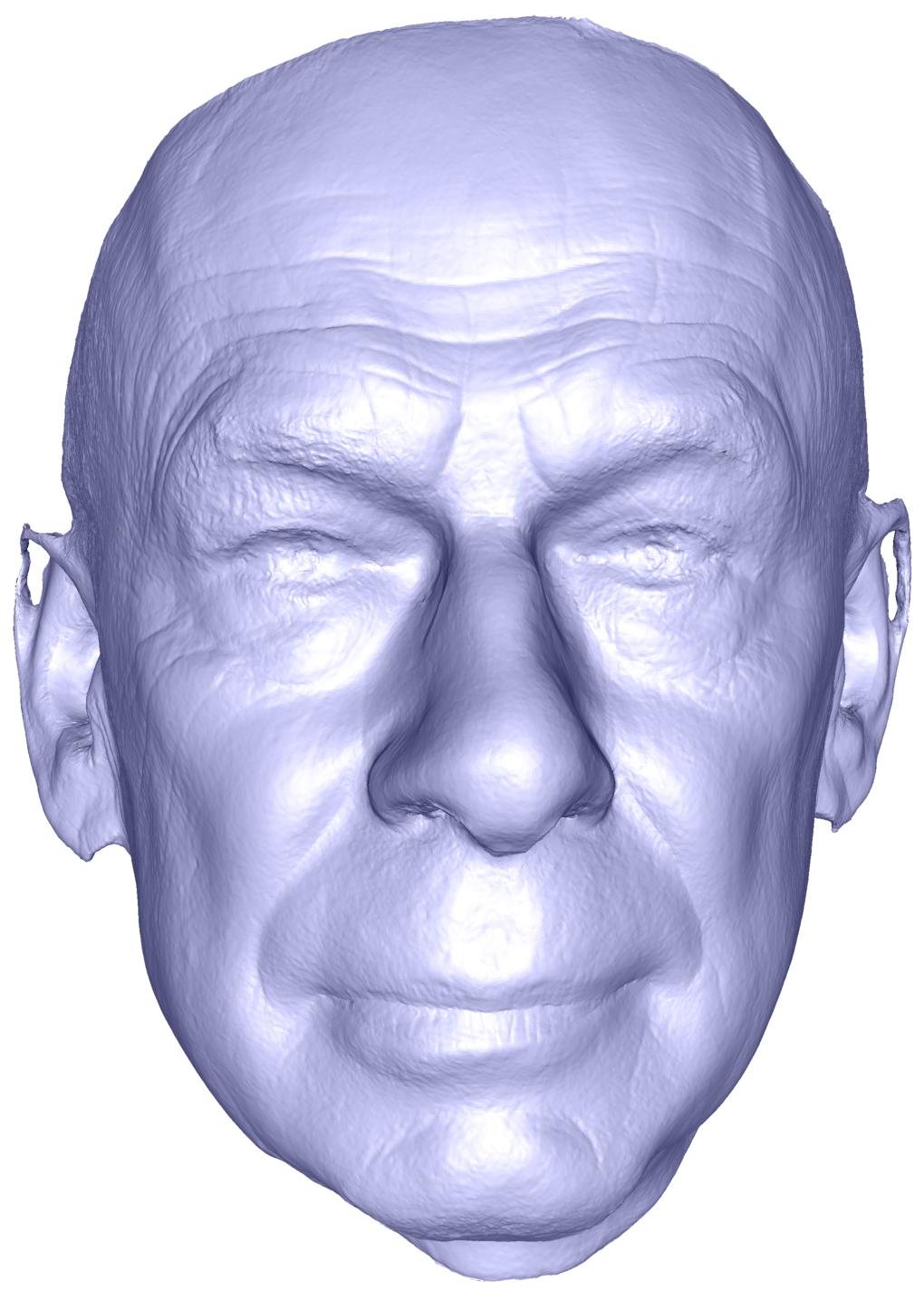} &
\includegraphics[width=4cm]{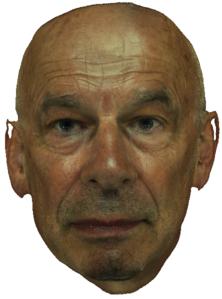}  \\
\end{tabular}

\caption{Geometry mesh and renderings of different subjects}
\label{fig:range_faces}
\end{figure*}

Finally, we present rendering results to demonstrate
the quality of our captured face models. We render
using the Cook Torrance model  \cite{CT.1982} and the hybrid
normals technique proposed by Ma et al. \cite{ma2007rapid} where
the estimated specular and diffuse surface normals
are used to shade respectively the specular and diffuse components of the BRDF. We use two slopes of
Beckmann functions to model the micro-facets distribution. In this work we assume constant roughness
parameters and refraction index across the face. In
Figure \ref{fig:rendering_regions} we show renderings of face details which
highlight the successful capture of face microgeometry along with reflectance properties necessary for
a photorealistic effect. In Figure \ref{fig:range_faces} we show the
captured geometry and renderings using the captured
reflectance properties for a range of faces. In spite of
using a simple reflectance model and making strong
assumptions about specular parameters being fixed
over the face surface, we are still able to achieve
highly realistic appearance. Our approach is able to
cope with facial hair.

\section{Conclusions}

In this work we present a practical 3D face acquisition approach that allows the capture of an ear-to-ear mesh along with the skin micro-geometry and reflectance properties. Our system requires no prior geometrical calibration. The cameras parameters are obtained by structure-from-motion and are used to estimate a base mesh which is further refined using the recovered photometric surface normals. To achieve an ear-to-ear coverage of the face and overcome the problem of fixed photometric viewpoint inherent in polarized spherical gradient illumination, we capture the face in three poses and robustly stitch the corresponding normal maps and intrinsic textures into a seamless, complete and detailed face model.

While providing a practical way to bypass the view-dependency issue inherent to polarized spherical gradient illumination, our multi-pose approach requires capturing the subject in three different poses which slightly lengthens the capture process and could be a handicap to tasks like expression or performance captures.

We aim at tackling this issue in future work by augmenting our setup with two additional photometric views matching the two profile poses. At each vertex of the dome, the number of polarized light can be locally multiplied such that the incident illumination from each vertex can be produced independently by more than one source. This would be different to the longitude/latitude approach proposed by Ghosh \etal \cite{ghosh2011multiview} in the fact that, instead of using a locally orthogonal polarisation pattern, we aim at simply assigning to each view an independent group of polarized lights. This will allow exact separation of diffuse and specular reflections to be retained. Such an approach will also enable multiview photometric constraints to be exploited in the shape reconstruction process.

\ifCLASSOPTIONcompsoc
  \section*{Acknowledgments}
\else
  \section*{Acknowledgment}
\fi

We are grateful to Hadi Dahlan for assistance with data collection and to Fufu Fang for assistance with lightstage calibration and programming.

\ifCLASSOPTIONcaptionsoff
  \newpage
\fi

\bibliographystyle{IEEEtran}
\bibliography{willbib,biblio,rtpc}

\begin{IEEEbiography}[{\includegraphics[width=1in,height=1.25in,clip,keepaspectratio]{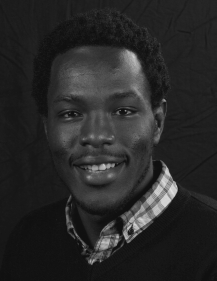}}]{Alassane Seck} received an M.sc. in Artificial Intelligence (Language and Image Processing) from the University of Caen, France and an M.sc in Image Processing and Remote Sensing from ENSEGID, Bordeaux, France. He is currently a PhD student in Computer Vision at Aberystwyth University. His research interests include Computer Vision, Computer Graphics, 3D surface texture modelling, 3D surface micro-structure measurements and Machine Learning.
\end{IEEEbiography}
\begin{IEEEbiography}[{\includegraphics[width=1in,height=1.25in,clip,keepaspectratio]{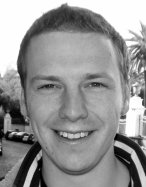}}]{William A. P. Smith} (M'08) received the B.Sc. degree in computer science, and the Ph.D. degree in computer vision from the University of York, York, U.K. He is currently a Senior Lecturer with the Department of Computer Science, University of York, York, U.K. His research interests are in face modeling, shape-from-shading, reflectance analysis and the psychophysics of shape-from-X. He has published more than 80 papers in international conferences and journals, was awarded the Siemens best security paper prize at BMVC 2007, and was finalist as the U.K. nominee for the ERCIM Cor Baayen award 2009. He is an associate editor of the IET journal Computer Vision.
\end{IEEEbiography}
\begin{IEEEbiography}[{\includegraphics[width=1in,height=1.25in,clip,keepaspectratio]{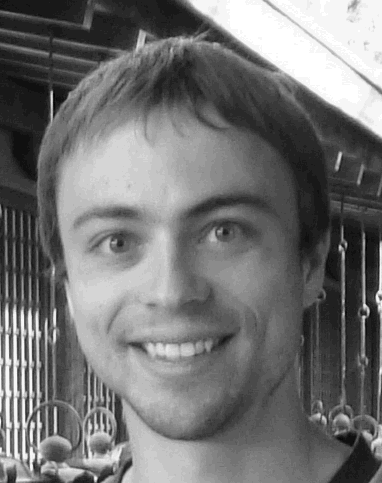}}]{Arnaud Dessein} received the Dipl.-Ing. degree from {\'{E}}cole Centrale de Lille, Lille, Fran\-ce, the M.Sc. degree in acoustics, signal processing and computer science applied to music, and the Ph.D. degree in computer science from Universit{\'{e}} Pierre et Marie Curie, Paris, France. He is currently a Research Associate with the Department of Computer Science, University of York, York, U.K. His research interests include computer vision, audio analysis, signal processing, machine learning, statistics and information theory. He has authored 1 book chapter, 2 international journal articles and 5 refereed conference papers. He is a member of SEE.
\end{IEEEbiography}
\begin{IEEEbiography}[{\includegraphics[width=1in,height=1.25in,clip,keepaspectratio]{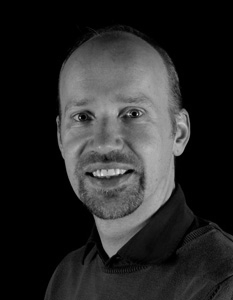}}]{Bernie Tiddeman} is a Senior Lecturer and Head of Department in Computer Science at Aberystwyth University.  He obtained his BSc from University of St Andrews in 1992, MSc from Manchester University in 1994 and PhD from Heriot-Watt University in 1998. From 1999-2010 he worked as a researcher and then lecturer at the University of St Andrews.  His research interests include 2D and 3D facial image analysis and synthesis, including texture modelling for age estimation and age progression and skin health analysis and synthesis.
\end{IEEEbiography}
\begin{IEEEbiography}[{\includegraphics[width=1in,height=1.25in,clip,keepaspectratio]{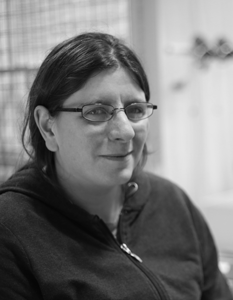}}]{Hannah M. Dee} is a Senior Lecturer in Computer Science at Aberystwyth University in the UK. Before this, she carried out post-doctoral work at the Grenoble Institute of Technology (INPG), the University of Leeds, and Kingston University. She received her Ph.D. in computer vision from the University of Leeds in 2006. Her research area is computer vision, with a particular interest in vision for modelling change, growth and texture. She is member of BCSWomen, the British Computer Society's group for women in technology, and is active in encouraging women and girls to consider careers in computer science.
\end{IEEEbiography}
\begin{IEEEbiography}[{\includegraphics[width=1in,height=1.25in,clip,keepaspectratio]{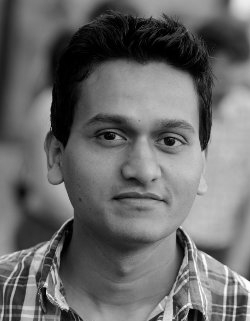}}]{Abhishek Dutta} received the Bachelor's degree in Computer Engineering from Tribhuvan University (Nepal) in 2009, the Master of Science (by research) in Computer Science from the University of York (UK) in 2012 and the PhD degree in Computer Science from the University of Twente (Netherlands) in 2015. He is now a Research Fellow in the Visual Geometry Group (VGG) of the Department of Engineering Sciences at Oxford University. His research interests are in the area of Computer Vision, Computer Graphics and Machine Learning.
\end{IEEEbiography}

\vfill

\end{document}